\documentclass[12pt]{article}
\usepackage{amsmath}
\usepackage{times}
\usepackage{graphicx}
\usepackage{color}
\usepackage{multirow}
\usepackage[authoryear]{natbib}
\usepackage{rotating}
\usepackage{bbm}
\usepackage{latexsym}
\usepackage{listings}
\usepackage{amssymb}

\newcommand{\captionfonts}{\normalsize}

\makeatletter  
\long\def\@makecaption#1#2{%
  \vskip\abovecaptionskip
  \sbox\@tempboxa{{\captionfonts #1: #2}}%
  \ifdim \wd\@tempboxa >\hsize
    {\captionfonts #1: #2\par}
  \else
    \hbox to\hsize{\hfil\box\@tempboxa\hfil}%
  \fi
  \vskip\belowcaptionskip}
\makeatother   
\begin{document}
\hspace{13.9cm}1

\ \vspace{10mm}\\

\noindent
{\LARGE \textbf{Theory of the superposition principle for randomized connectionist representations in neural networks}}

\ \\
{\bf \large E. Paxon Frady$^{\displaystyle 1}$,}
{\bf \large Denis Kleyko$^{\displaystyle 2}$,}
{\bf \large Friedrich T. Sommer$^{\displaystyle 1}$}\\
{$^{\displaystyle 1}$Redwood Center for Theoretical Neuroscience, U.C. Berkeley}\\
{$^{\displaystyle 2}$Department of Computer Science, Electrical and Space Engineering, Lulea University of Technology}\\

%

%\ \\[-2mm]

{\bf Running Title:} Theory of superposition retrieval and capacity
\thispagestyle{empty}
%\markboth{}{NC instructions}
%%
%\ \vspace{-0mm}\\
%%
%%Abstract
\begin{center} {\bf Abstract} \end{center}
To understand cognitive reasoning in the brain, it has been proposed that symbols and compositions of symbols are represented by activity patterns (vectors) in a large population of neurons. Formal models implementing this idea \citep{Plate2003, Kanerva2009, Gayler2003, Eliasmith2012} include a reversible superposition operation for representing with a single vector an entire set of symbols or an ordered sequence of symbols. If the representation space is high-dimensional, large sets of symbols can be superposed and individually retrieved. However, crosstalk noise limits the accuracy of retrieval and information capacity. To understand information processing in the brain and to design artificial neural systems for cognitive reasoning, a theory of this superposition operation is essential. Here, such a theory is presented. The superposition operations in different existing models are mapped to linear neural networks with unitary recurrent matrices, in which retrieval accuracy can be analyzed by a single equation.  We show that networks representing information in superposition can achieve a channel capacity of about half a bit per neuron, a significant fraction of the total available entropy.  
Going beyond existing models, superposition operations with recency effects are proposed that avoid catastrophic forgetting when representing the history of infinite data streams. These novel models correspond to recurrent networks with non-unitary matrices or with nonlinear neurons, and can be analyzed and optimized with an extension of our theory.

%%%%%%%%%%%

\newpage

\tableofcontents

\newpage

\section{Introduction}

Various models of cognitive reasoning have been proposed that use high-dimensional vector spaces to represent symbols, items, tokens or concepts. Such models include holographic reduced representations (HRR) \citep{Plate2003}, and hyperdimensional computing (HDC) \citep{Kanerva2009}. In the following, these models will all be referred to by the term \emph{vector symbolic architectures} (VSA; see \citet{Gayler2003}; Methods \ref{sec:vsa_basics}). Recently, VSA models attracted significant novel interest 
for enhancing the functionality of standard neural networks to achieve complex cognitive and machine-learning tasks \citep{Eliasmith2012}, inductive reasoning \citep{Rasmussen2011}, and working memory \citep{Graves2014, Graves2016, Danihelka2016}.

VSA models use randomization to form distributed representations for atomic symbols and offer two distinct ways of combining symbols into new composite symbols, by superposition or binding. Superposition represents sets or sequences (ordered sets) of symbols by a single new vector, enabling each individual symbol to be accessed easily. For example, superposition can form a vector that makes a set of symbols quickly accessible that are relevant for an ongoing computation \citep{Rinkus2012, Kleyko2014}, akin to working memory in a brain, or the processor registers in a von Neumann computer. Second, superposition can be used to parallelize an operation by applying it once to a set of superposed vector symbols, rather than many times to each individual symbol. Conversely, binding serves to express specific relationships between pairs of symbols, for example, key-value bindings. 

Specific VSA models differ in how vector representations of atomic symbols are formed and how binding is implemented (see Results~\ref{sec:vsa_literature} and Methods \ref{sec:vsa_basics}). However, in all VSA models the superposition operation is lossy, and superposed elements are retrieved by measuring similarity between the superposition vector and individual code vectors. A symbol is contained in a superposition if the corresponding similarity is large. Such a superposition scheme works perfectly if code vectors are orthogonal, and VSA models exploit the fact that high-dimensional random vectors are \emph{pseudo-orthogonal}, that is, nearly orthogonal with high probability. However, some cross-talk noise is present with pseudo-orthogonal vectors, which limits retrieval accuracy.

In this study, we focus on the retrieval accuracy of a sequence of tokens represented in a single vector by superposition. To analyze the accuracy of retrieval, our study maps superposition and binding in existing VSA models to equivalent recurrent neural networks (RNN).   The resulting neural networks resemble models in the literature, such as temporal learning networks \citep{Williams1989cs}, echo-state machines \citep{Jaeger2002a, Lukosevicius2009}, liquid-state machines \citep{Maass2002}, and related network models \citep{Ganguli2008, Sussillo2009}. 
However, the RNNs that correspond to VSA models share properties that makes the analysis much simpler than for general recurrent networks. This enables us to derive a theory for recall accuracy and information capacity. 

The theory accurately predicts simulation experiments and reveals theoretical limits of performance that exceed those derived in earlier analyses \citep{Plate2003, Gallant2013}. Interestingly, our theory can explain why many VSA models from the literature exhibit the same retrieval accuracy and channel capacity of about 0.5 bits per neuron. We further found that a recurrent neural network, in which the superposition representation is optimized by learning, does not exceed the performance of existing VSA models. 
 
Finally, we investigate superposition operations with recency effect, in which inputs from the distant past are attenuated and gradually forgotten. We show that RNNs with either contracting recurrent weights or saturating non-linear activation functions have similar affect on information retrieval: the recency effect enables the network to form a \emph{memory buffer} of the recent symbols in an infinite sequence. The RNNs we propose generalize several VSA architectures which use a non-linear transfer function to constrain the activations of the memory vector \citep{Kanerva1996, Rachkovskij2001}. We derive the time constant and parameters that maximize the information content of such memory buffers, and illustrate trade-offs between recall accuracy of recent and distant inputs. 

With the theory presented here, neural networks based on VSAs can be appropriately scaled and optimized to handle communication or memory demands.

\section{Results}

\subsection{Superposing and retrieving a sequence of tokens in a RNN}

Assume a set of $D$ tokens (such as $D=27$ for letters of the alphabet and space) and a coding prescription of a VSA model which assigns an $N$-dimensional (such as $N=10,000$) random vector-symbol to represent each of the tokens. The representation can be summarized in the codebook matrix $\mathbf{\Phi} \in I\!\! R^N \times I\!\! R^D$. A sequence of $M$ items $\{ \mathbf{a}(m): m=1,..., M \}$ is the input, with $\mathbf{a}(m) \in I\!\! R^D$ a random one-hot or zero vector, represented as the vector-symbol $\mathbf{\Phi}_{d'} = \mathbf{\Phi}\mathbf{a}(m)$. Each item in the sequence can be encoded and stored into a single memory vector $\mathbf{x} \in I\!\!R^N$ by basic mechanisms in a recurrent neural network (RNN):
\begin{equation}
\mathbf{x}(m) = f( \mathbf{W} \mathbf{x}(m-1) + \mathbf{\Phi} \mathbf{a} (m) )
\label{eqn:rnn}
\end{equation}
\noindent
where $f(\mathbf{x})$ is the neural activation function -- executed component-wise on the argument vector and $\mathbf{W} \in I\!\! R^N \times I\!\! R^N$ is the matrix of recurrent weights. We assume $\mathbf{W} = \lambda \mathbf{U}$ with $0 < \lambda \leq 1$ and $\mathbf{U}$ a unitary matrix, (i.e. $\mathbf{U}$ is an isometry with all $N$ eigenvalues equal to 1, $\mathbf{U} ^{\top} \mathbf{U} = \mathbf{I}$).    
The recurrent matrix can encapsulate the binding operation for different VSAs. As the input streams into the network, it is iteratively bound to different memory addresses by the recurrent matrix multiply each time step. This is known as trajectory-association \citep{Plate2003}. The instantaneous network activity state $\mathbf{x}(M)$ contains the history of the sequence in superposition, and the retrieval of an individual item $\mathbf{a} (M-K)$ at position $M-K$ in the sequence is done by:
\begin{equation}
\hat{\mathbf{a}}(M-K) = WTA \left( \mathbf{V}(K) \mathbf{x}(M) \right)
\label{eqn:readout}
\end{equation} 
where $\mathbf{V}(K) \in I\!\!R^D \times I\!\!R^N$ is a linear transform dereferencing the input that occurred $K$ time steps previously, and $WTA$ is the winner-take-all function returning a one-hot vector corresponding to the maximum of the argument. With this set up, the decoding matrix $\mathbf{V}(K)$ does not need to be learned, but can be computed from the recurrent weights by $\mathbf{V}(K) = \mathbf{\Phi}^\top \mathbf{W}^{-K}$.

\subsubsection{Retrieval accuracy of superposed random vectors}
\label{sec:capacity_theory}

The central piece in our theory is a formula for the retrieval accuracy -- the probability that (\ref{eqn:readout}) correctly identifies individual memory items that were input. To retrieve the item that was integrated $K$ time steps ago, the vector argument into the winner-take-all function is:
\begin{equation}
\begin{split}
    \mathbf{h}_{d} &:= \mathbf{V}_d(K) \mathbf{x}(M) = (\mathbf{\Phi}_{d})^\top \mathbf{W}^{-K} \mathbf{x}(M) 
\end{split}
\label{eqn:innerprod}
\end{equation}
with $\mathbf{\Phi}_d$ being the $d$-th column of the encoding matrix. If $d'$ is the index of the letter actually encoded $K$ time steps ago, then the readout is correct if $\mathbf{h}_{d'} > \mathbf{h}_{d}$ for all distractors $ d \not = d'$. Note, $\mathbf{\Phi}_{d'} = \mathbf{\Phi}\mathbf{a}(M-K)$.

The effect of one iteration of equation (\ref{eqn:rnn}) on the probability distribution of the network state $\mathbf{x}(m)$ is a Markov chain stochastic process, governed by the Chapman-Kolmogorov equation \citep{Papoulis1984}):
\begin{equation}
p(\mathbf{x}(m+1) | \mathbf{b}(m)) = \int  p(\mathbf{x}(m+1) | \mathbf{x}(m), \mathbf{b}(m)) \; p(\mathbf{x}(m)) \; d\mathbf{x}(m)
\label{eqn:ptpo}
\end{equation}
with $\mathbf{b}(m) := \mathbf{\Phi} \; \mathbf{a}(m)$ describing the input symbol and the transition kernel:
\begin{equation}
p(\mathbf{x}(m+1) | \mathbf{x}(m), \mathbf{b}(m)) = \delta(\mathbf{x}(m) - f(\mathbf{W} \mathbf{x}(m) + \mathbf{b}(m)))
\label{eqn:det_sys}
\end{equation}

Thus to compute the distribution of (\ref{eqn:innerprod}) in the general case, one has to iterate equations (\ref{eqn:ptpo}) and (\ref{eqn:det_sys}).

The situation simplifies considerably if the input sequence to be memorized is encoded according to one of the VSA methods. Each VSA method maps to a particular encoding matrix $\mathbf{\Phi}$ and recurrent matrix $\mathbf{W}$ that satisfy the following conditions:
\begin{itemize}
\item Code vectors $\mathbf{\Phi}_d$ are composed from \emph{identically distributed} components: 
\begin{equation}
%p((\mathbf{\Phi}_{d'})_i) = p((\mathbf{\Phi}_{d})_j) \;\forall j, d
p((\mathbf{\Phi}_{d})_i) \sim p_{\mathbf{\Phi}}(x) \;\forall i, d
\label{eqn:phiiid}
\end{equation}
where $p_{\mathbf{\Phi}}(x)$ is the distribution for a single component of a random code vector. 

\item Components within a code vector and between code vectors are \emph{independent}:
\begin{equation}
 p \left( (\mathbf{\Phi}_{d'})_i, (\mathbf{\Phi}_{d})_j \right) = p((\mathbf{\Phi}_{d'})_i) \; p((\mathbf{\Phi}_{d})_j) \;\forall j\not = i \vee d' \not = d
\label{eqn:phiind}
\end{equation}

\item The recurrent weight matrix $\mathbf{W}$ \emph{preserves mean and variance} of every component of a code vector:
\begin{align}
\label{eqn:wiid}
\begin{split}
E((\mathbf{W} \mathbf{\Phi}_{d})_i) &= E((\mathbf{\Phi}_d)_i)  \;\; \forall i, d
\\
\mbox{Var}((\mathbf{W} \mathbf{\Phi}_{d})_i) &= \mbox{Var}((\mathbf{\Phi}_d)_i)  \;\; \forall i, d
\end{split}
\end{align}
 
\item The recurrent matrix preserves independence with a \emph{large cycle time}:
\begin{equation}
p((\mathbf{W}^{m}\mathbf{\Phi}_{d})_i, (\mathbf{\Phi}_{d})_i)  = p((\mathbf{W}^{m}\mathbf{\Phi}_{d})_i) \; p((\mathbf{\Phi}_d)_i)  \;\; \forall i, d; m = \{1, ... , O(N) \}
\label{eqn:wind}
\end{equation}
\end{itemize}

Under the conditions (\ref{eqn:phiiid})-(\ref{eqn:wind}), expression (\ref{eqn:innerprod}) becomes a sum of $N$ independent random variables: 
\begin{equation}
\mathbf{h}_d = \sum_{i=1}^N \mathbf{V}_{d, i}(K) \mathbf{x}_i(M) =
\sum_{i=1}^N (\mathbf{\Phi}_{d})_i (\mathbf{W}^{-K} \mathbf{x}(M))_i =
\sum_{i=1}^N \mathbf{z}_{d, i}
\label{eqn:h_cl}
\end{equation}

For large enough $N$, the central limit theorem holds and the distribution of $\mathbf{h}_d$ can be well approximated by a Gaussian: $p(\mathbf{h}_d) \sim \mathcal{N}(N\mu_d, N \sigma_d^2)$ with $\mu_d := E(\mathbf{z}_{d,i})$ and $\sigma^2_d := Var(\mathbf{z}_{d,i})$ the mean and variance of $\mathbf{z}_{d,i}$.

There are two distinct types of retrieval we will analyze: \emph{classification} and \emph{detection}. For classification, each position in the sequence contains a symbol and the task during retrieval is to retrieve the identity of the symbol. For detection, some positions in the sequence can be empty, corresponding to 
inputs where $\mathbf{a}(m)$ is a zero vector. The retrieval task is then to detect whether or not a symbol is present, and, if so, reveal its identity. Thus, detection requires a rejection threshold, $\theta$, which governs the trade-off between the two error types: \emph{misses} and \emph{wrong rejections}.

For \emph{classification}, the retrieval is correct if $\mathbf{h}_{d'}$ is larger than $\mathbf{h}_{d}$ for all the $D-1$ distractors $d\not = d'$. The \emph{classification accuracy}, $p_{corr}$, is:  
\begin{align}
\begin{split}
p_{corr}(K) &= p\left(\mathbf{h}_{d'} > \mathbf{h}_{d} \; \forall d \not = d' \right) \\
&= \int_{-\infty}^{\infty} p(\mathbf{h}_{d'}=h)\left[p(\mathbf{h}_{d} < h) \right]^{D-1} dh \\
&= \int_{-\infty}^{\infty} \mathcal{N}(h;N\mu_{d'}, N\sigma_{d'}^2) \left[ \int_{-\infty}^{h} \mathcal{N} (h'; N\mu_d, N\sigma_d^2) \;dh' \right]^{D-1} dh
\end{split}
\label{eqn:p_corr}
\end{align}
For clarity, the argument variable is added in the notation of the Gaussian distributions, $p(x) = \mathcal{N} (x; \mu, \sigma^2)$. 

The Gaussian variables $h$ and $h'$ in (\ref{eqn:p_corr}) can be shifted and rescaled to yield:
\begin{equation}
p_{corr}(K) = \int_{-\infty}^{\infty} \frac{dh}{\sqrt{2\pi}} \  e ^ {- \frac{1}{2} h^2} \left[ \Phi \left( \frac{\sigma_d}{\sigma_{d'}}h - \sqrt{N}\;\frac{ \mu_{d}-\mu_{d'}}{\sigma_{d'}} \right) \right] ^ {D-1} 
\label{eqn:p_corr_full}
\end{equation}
where $\Phi$ is the Normal cumulative density function. Further simplification can be made when $\sigma_{d'} \approx \sigma_d$:
\begin{equation}
p_{corr}(K) =  \int_{-\infty}^{\infty} \frac{dh}{\sqrt{2\pi}} \  e ^ {- \frac{1}{2} h^2} \left[ \Phi \left( h + s(K) \right) \right] ^ {D-1}
\label{eqn:p_corr_s}
\end{equation}
with the {\it signal-to-noise ratio} (SNR), $s$, defined as:
\begin{equation}
 s(K) := \sqrt{N}\;\frac{\mu_{d'} - \mu_{d}} {\sigma_{d}} 
 \label{eqn:snr}
 \end{equation}

As a first sanity check, consider (\ref{eqn:p_corr_s}) in the vanishing signal-to-noise regime, that is for $s\to 0$. The first factor in the integral, the Gaussian, becomes then the inner derivative of the second factor, the cumulative Gaussian raised to the $(D-1)$th power. With $s \to 0$, the integral can be solved analytically using the (inverse) chain rule to yield the correct chance value for the classification: $p_{corr} = \frac{1}{D} \Phi(h)^D|_{-\infty}^{\infty} = \frac{1}{D}$. 

For \emph{detection}, the retrieval task is to detect whether or not an input item was integrated $K$ time steps ago, and, if so, to identify the input token. This type of input stream has been denoted as a \emph{sparse input sequence} \citep{Ganguli2010}. In this case, if none of the $\mathbf{h}_{d}$ variables exceed a detection threshold $\theta$, then the network will output that no item was stored. Analogous to equation (\ref{eqn:p_corr_s}), the probability of correct decoding during detection, the {\it detection accuracy} is:
\begin{align}
\begin{split}
p^\theta_{corr}(K) &= p\left((\mathbf{h}_{d'} > \mathbf{h}_{d} \;\forall d \not = d')\wedge (\mathbf{h}_{d'} \geq \theta) \right) \\
&= \int_{\theta}^{\infty} \frac{dh}{\sqrt{2 \pi}} e^{-\frac{1}{2}h^2} \left[ \Phi\left(h + s(K) \right) \right]^{D-1}   
\end{split}
 \label{eqn:p_corr_rej}
\end{align}

Note that (\ref{eqn:p_corr_rej}) is of the same form as (\ref{eqn:p_corr_s}), but with different integration bounds. In particular, $p_{corr}(K) \geq p^\theta_{corr}(K)$, that is, the accuracy is bigger for recognition than for detection, since detection can fail because of two types of errors, detection errors and misclassification.  

Equations (\ref{eqn:p_corr_s}) and (\ref{eqn:p_corr_rej}) will be exploited to analyze various different coding schemes (Results \ref{sec:vsa}) and network nonlinearities (Results \ref{sec:recency}).

\subsubsection{The regime of high-fidelity retrieval}
\label{sec:p_corr_approx}

\begin{figure}[t]
\centering
\includegraphics[width=0.8\textwidth]{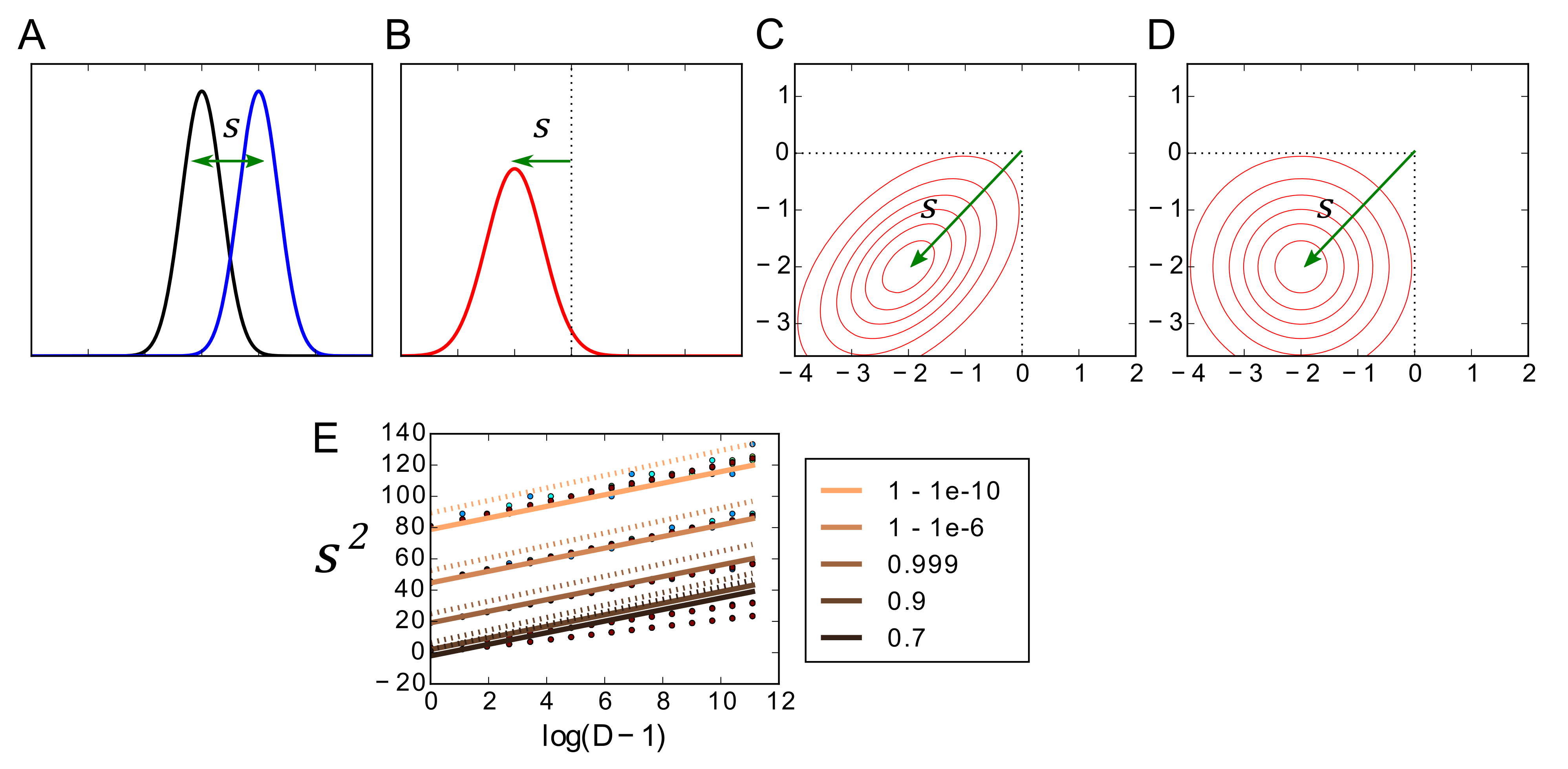}
\caption[High-fidelity approximation of $p_{corr}$]{\textbf{Approximating the retrieval accuracy in the high-fidelity regime.} A. The retrieval is correct when the value drawn from distribution $p(\mathbf{h}_{d'})$ (blue) exceeds the values produced by  $D-1$ draws from the distribution $p(\mathbf{h}_{d})$ (black). In the shown example the signal-to-noise ratio is $s=2$. B. When $D=2$, the two distributions can be transformed into one distribution describing the difference of both quantities, $p(\mathbf{h}_{d'}-\mathbf{h}_{d})$. C. When $D>2$, the $D-1$ random variables formed by such differences are correlated. Thus, in general, the multivariate cumulative Gaussian integral ($\ref{eqn:plfdD}$) cannot be factorized. Example shows the case $D=3$, the integration boundaries displayed by dashed lines. D. However, for large $s$, that is, in the high-fidelity regime, the factorial approximation (\ref{eqn:factorialapro}) becomes quite accurate. Panel shows again the $D=3$ example. 
E. Linear relationship between the parameters $s^2$ and the logarithm of $D$. The numerically evaluated full theory (dots) coincides more precisely with the approximated linear theories (lines) when the accuracy is high (accuracy indicated by copper colored lines; legend). The simpler linear theory (\ref{eqn:nm_chernoff}; dashed lines) matches the slope of the full theory but exhibits a small offset. The more elaborate linear theory (\ref{eqn:p_corr_chang}; solid lines) provides a quite accurate fit of the full theory for high accuracy values.}
\label{fig:p_corr_approx}
\end{figure}

Consider the special case $D=2$. Since the (rescaled and translated) random variables $p(\mathbf{h}_{d'}) \sim \mathcal{N}(s, 1)$ and $p(\mathbf{h}_d) \sim \mathcal{N}(0, 1)$ (Fig. \ref{fig:p_corr_approx}A) are uncorrelated, one can switch to a new Gaussian variable representing their difference: $\mathbf{y}:= \mathbf{h}_d-\mathbf{h}_{d'}$ with $p(\mathbf{y}) \sim {\cal N}(-s, 2)$ (Fig. \ref{fig:p_corr_approx}B). Thus, for $D=2$ one can compute (\ref{eqn:p_corr_s}) by just the Normal cumulative density function (and avoiding the integration):
\begin{equation}
p_{corr}(K; D=2) = p(\mathbf{y} < 0) =  \Phi\left(\frac{s}{\sqrt{2}}\right)
 \label{eqn:plfd2}
\end{equation}
The result (\ref{eqn:plfd2}) is the special case $d=1$ of table entry ``$10,010.8$'' in Owen's table of normal integrals \citep{owen1980table}.

In general, for any $D>2$, the $p_{corr}$ integral cannot be solved analytically, but can be numerically approximated to arbitrary precision (Methods Fig. \ref{fig:p_corr_algo}). Next, we derive steps to approximate the accuracy in the high-fidelity regime, analogous to previous work in VSA models  \citep{Plate2003, Gallant2013} showing that accuracy of retrieval scales linearly with the dimensionality of the network ($N$). We will compare the tightness of the approximations we derive here with these previous results.

Let's now try to apply to the case $D>2$ what worked for $D=2$ (\ref{eqn:plfd2}), that is, get rid of the integral in (\ref{eqn:p_corr_s}) by transforming to new variables $\mathbf{y}_d=\mathbf{h}_d-\mathbf{h}_{d'}$ for each of the $D-1$ distractors with $d \ne d'$. The difficulty with more than one of the $\mathbf{y}_d$ variables is that they are not uncorrelated: $\mbox{Cov}(\mathbf{y}_i \mathbf{y}_j) = E((\mathbf{y}_i-s)(\mathbf{y}_j-s)) = 1$. 
Thus, $p(\mathbf{y}) \sim {\cal N}(-\mathbf{s}, \mathbf{\Sigma}^2)$ with:
\begin{equation*}
\mathbf{s} =  \left( \begin{array}{c} 
s\\
s\\
...\end{array} \right) \in I\!\!R^{D-1}   
, \; 
\mathbf{\Sigma}^2 = \left( \begin{array}{ccc}
2 & 1 & 1 \\
1 & 2 & ... \\
1 & ... & 2 \end{array} \right) \in I\!\!R^{D-1} \!\!\times I\!\!R^{D-1}
 \label{ssigma}
\end{equation*}

In analogy to the $D=2$ case we can now compute the result of (\ref{eqn:p_corr_s}) for $D>2$ by:
\begin{equation}
p_{corr}  =   p(\mathbf{y}_d < 0 \; \forall d \ne d') =  \Phi_{D-1}(\mathbf{s}, \mathbf{\Sigma}^2)
 \label{eqn:plfdD}
\end{equation}
with multivariate cumulative Gaussian, $\Phi_{D-1}(\mathbf{s}, \mathbf{\Sigma}^2)$, using table entry ``$n0,010.1$'' in Owen's table of normal integrals \citep{owen1980table}.

For uncorrelated variables, the covariance matrix $\mathbf{\Sigma}^2$ is diagonal and the multivariate cumulative distribution factorizes. 
However, in our case the covariance has positive uniform entries in the off-diagonal, which means that the $\mathbf{y}_d$'s are uniformly correlated and lie on an ellipse aligned with $(1, 1, ...)$ (Fig. \ref{fig:p_corr_approx}C). 
In the high signal-to-noise regime, the integration boundaries are removed from the mean and the exact shape of the distribution should not matter so much (Fig. \ref{fig:p_corr_approx}D). Thus, the first step takes the \emph{factorized approximation} (FA) to the multivariate Gaussian to approximate $p_{corr}$ in the high signal-to-noise regime:
\begin{equation}
\begin{split}\
p_{corr:\ FA}  &= \left[ \Phi \left( \frac{s}{\sqrt{2}} \right) \right]^{D-1}
\end{split}
\label{eqn:factorialapro}
\end{equation} 
Note that for $s \to 0$ in the low-fidelity regime, this approximation is not tight; the chance probability is $1/D$ when $s \to 0$, but equation (\ref{eqn:factorialapro}) yields $0.5^{D-1}$ which is way too low for $D>2$.

To obtain an analytic expression, an approximate formula for the one-dimensional cumulative Gaussian, which is closely related to the complementary error function, $\Phi(x) = 1 - \frac{1}{2} \mbox{erfc}(x/\sqrt{2})$, is needed.
There is a well-known exponential upper bound on the complementary error function, the \textit{Chernoff-Rubin bound} (CR) \citep{chernoff1952measure}, improved by \citep{jacobs1966probability, bound1970probability}: $\mbox{erfc}(x) \leq B_{CR}(x) = e^{-x^2}$. Using  $x= s/\sqrt{2}$, we obtain $B_{CR}(x/\sqrt{2}) = e^{-s^2/4}$, which can be inserted into (\ref{eqn:factorialapro}) as the next step to yield an approximation of $p_{corr}$:
\begin{equation}
p_{corr:\ FA-CR} = \left[1- \frac{1}{2} e^{-s^2/4} \right]^{D-1}  
\label{eqn:p_corr_facr}
\end{equation}

The final step, using the \textit{local error expansion} $e^x = 1 + x + \dots $ when $x$ is near 0 (LEE), we can set $x = -\frac{1}{2}e^{-s^2/4}$ and rewrite:
\begin{equation}
p_{corr:\ FA-CR-LEE} =  1 - \frac{1}{2} (D-1) e^{-s^2/4}
\label{eqn:p_corr_apx}
\end{equation}
Solving for $s^2$ gives us a simple relationship:
 \begin{equation}
s^2 = 4 \left[ \ln(D-1) - \ln (2 \epsilon) \right]
\label{eqn:nm_chernoff}
 \end{equation}
where $\epsilon := 1 - p_{corr}$.
 
The approximation (\ref{eqn:nm_chernoff}) is quite accurate (Fig. \ref{fig:p_corr_approx}E, dashed lines) but not tight. Even if (\ref{eqn:factorialapro}) was tight in the high-fidelity regime, there would still be a discrepancy because the CR bound is not tight. This has been noted for long time, enticing varied efforts to improve the CR bound, usually involving more complicated multi-term expressions, e.g., \citep{chiani2003new}. Quite recently, \citet{chang2011chernoff} studied one-term bounds of the complementary error function of the form $B(x; \alpha, \beta): = \alpha e^{-\beta x^2}$. First, they proved that there exists no parameter setting for tightening the original Chernoff-Rubin upper bound. Second, they reported a parameter range where the one-term expression becomes a lower bound:  $\mbox{erfc}(x) \geq B(x; \alpha, \beta)$ for $x \geq 0$. The lower bound becomes the tightest for with $\beta = 1.08$ and $\alpha=\sqrt{\frac{2e}{\pi}}\frac{\sqrt{\beta-1}}{\beta}$. This setting approximates the complementary error function as well as an 8-term expression derived in \citet{chiani2003new}.  
Following \citet{chang2011chernoff}, we approximate the cumulative Gaussian with the \textit{Chang} bound (Ch), and follow the same FA and LEE steps to derive a tighter linear fit to the true numerically evaluated integral:
 \begin{equation}
  s^2 = \frac{4}{\beta} \left[ \ln(D-1) - \ln(2\epsilon) + \ln\left(\sqrt{\frac{2e}{\pi}} \frac{\sqrt{\beta-1}}{\beta}\right) \right] 
 \label{eqn:p_corr_chang}
 \end{equation}
 with $\beta = 1.08$. This fits the numerically evaluated integral in the high-fidelity regime (Fig. \ref{fig:p_corr_approx}E, solid lines), but is not as accurate in the low-fidelity regime.

\subsubsection{Channel capacity of a superposition}

The channel capacity \citep{Feinstein1954} of a superposition can now be defined as the maximum of the mutual information between the true sequence and the sequence retrieved from the superposition state $\mathbf{x}(M)$. The mutual information between the individual item that was stored $K$ time steps ago ($\mathbf{a}_{d'}$) and the item that was retrieved ($\hat{\mathbf{a}}_d$) is given by:
\begin{equation*}
I_{item} = \sum_{d}^D \sum_{d'}^D p(\hat{\mathbf{a}}_d, \mathbf{a}_{d'}) \log_2 \left( \frac {p(\hat{\mathbf{a}}_d, \mathbf{a}_{d'}) }{p(\hat{\mathbf{a}}_d) p(\mathbf{a}_{d'})} \right)
\end{equation*}

Because the sequence items are chosen uniformly random from the set of tokens, both the probability of a particular token as input and the probability of a particular token as the retrieved output are the same: $p(\hat{\mathbf{a}}_d) = p(\mathbf{a}_{d'}) = 1/D$. The $p_{corr} (K)$ integral evaluates the conditional probability that the output item is the same as the input item:
\begin{equation*}
 p(\hat{\mathbf{a}}_{d'} | \mathbf{a}_{d'}) = \frac{p(\hat{\mathbf{a}}_{d'}, \mathbf{a}_{d'})}{p(\mathbf{a}_{d'})} = p_{corr} (K)
\end{equation*}

To evaluate the $p(\hat{\mathbf{a}}_d, \mathbf{a}_{d'}) \  \forall d \ne d'$ terms, $p_{corr} (K)$ is needed to compute the probability of choosing the incorrect token given the true input. The probability that the token is retrieved incorrectly is $1 - p_{corr}(K)$, and each of the $D-1$ distractors is equally likely to be the incorrectly retrieved token, thus:
\begin{equation*}
p(\hat{\mathbf{a}}_d | \mathbf{a}_{d'}) = \frac{p(\hat{\mathbf{a}}_d, \mathbf{a}_{d'})}{p(\mathbf{a}_{d'})} = \frac{1-p_{corr}(K)}{D-1} \ \ \ \forall \  d \ne d'
\end{equation*} 
Plugging these into the mutual information and simplifying:
\begin{align}
\begin{split}
I_{item}(p_{corr}(K)) &= p_{corr}(K) \log_2 \left( p_{corr}(K) D \right) \\
&+ (1-p_{corr}(K)) \log_2 \left( \frac{D}{D-1} \left( 1 - p_{corr} (K) \right) \right)
\end{split}
\label{eqn:info_item}
\end{align}

The total mutual information is then the sum of the information for each item in the full sequence:
\begin{equation}
%I_{total} = \sum_{K=1}^M \left[ p_{corr}(K) \log_2 \left( p_{corr}(K) D \right) + (1 - p_{corr}(K)) \log_2 \left( \frac{D}{D-1} \left( 1 - p_{corr}(K) \right) \right) \right]
I_{total} = \sum_{K=1}^M I_{item}(p_{corr}(K))
\label{eqn:total_infoK}
\end{equation}

Note that if the accuracy is the same for all items: $I_{total} = M\; I_{item}(p_{corr})$, and by setting $p_{corr} = 1$ one obtains the entire information entered into the hypervector: $I_{stored} = M \log_2(D)$.

The channel capacity per neuron is then $I_{total}/N$  (\ref{eqn:total_infoK}) using the parameter settings which maximize this quantity. We next analyze different VSA schemes in order to estimate memory accuracy $p_{corr}$, and channel capacity.

\subsection{VSA models that correspond to linear RNNs} 
\label{sec:vsa}

We begin by analyzing the VSA models that can be mapped to an RNN with linear neurons, $f(\mathbf{x}) = \mathbf{x}$. In this case, a sequence of inputs $\{ \mathbf{a}(1), ..., \mathbf{a}(M) \}$ into the RNN (\ref{eqn:rnn}) produce the memory vector:
\begin{equation}
    \mathbf{x}(M) = \sum_{m=1}^{M} \mathbf{W}^{M-m} \mathbf{\Phi} \mathbf{a}(m)
    \label{eqn:bsc_traj}
\end{equation}
and we can compute $\mathbf{z}_{d,i}$ from (\ref{eqn:h_cl}):
\begin{align}
\begin{split}
\mathbf{z}_{d,i} &= (\mathbf{\Phi}_d)_i (\mathbf{W}^{-K} \mathbf{x}(M))_i \\
&= 
\begin{cases} 
(\mathbf{\Phi}_{d'})_i (\mathbf{\Phi}_{d'})_i + \sum_{m \ne (M-K)}^M (\mathbf{\Phi}_{d'})_i (\mathbf{W}^{M-K-m} \mathbf{\Phi}_{d'})_i & \mbox{ if } d=d' \\
\sum_{m}^M (\mathbf{\Phi}_{d})_i (\mathbf{W}^{M-K-m} \mathbf{\Phi}_{d'})_i & \mbox{ otherwise }
\end{cases}
\end{split}
\label{eqn:z_stats}
\end{align}

Given the conditions (\ref{eqn:phiiid})-(\ref{eqn:wind}), the statistics of the codewords are preserved by the recurrent matrix, which must be unitary (i.e. $p(\mathbf{W}^{m}\mathbf{\Phi}_{d}) \sim p_{\mathbf{\Phi}}(x) \ \forall m = \{1, ..., M\}$). The mean and variance of $\mathbf{z}_{d,i}$ is:
\begin{equation}
\mu_d = 
 \begin{cases}  E_{\mathbf{\Phi}}(x^2) + (M-1) E_{\mathbf{\Phi}}(x)^2  & \mbox{ if } d=d' \\
M E_{\mathbf{\Phi}}(x)^2  & \mbox{ otherwise }
\end{cases}
\label{eqn:z_mean}
\end{equation}
\begin{equation}
\sigma^2_d = 
 \begin{cases}  V_{\mathbf{\Phi}}(x^2) + (M-1) V_{\mathbf{\Phi}}(x)^2  & \mbox{ if } d=d' \\
M V_{\mathbf{\Phi}}(x)^2  & \mbox{ otherwise }
\end{cases}
\label{eqn:z_var}
\end{equation}
with $E_{\mathbf{\Phi}}(g(x)):= \int g(x) p_{\mathbf{\Phi}}(x) dx$, $V_{\mathbf{\Phi}}(g(x)):= E_{\mathbf{\Phi}}(g(x)^2) - E_{\mathbf{\Phi}}(g(x))^2$ being the mean and variance of $p_{\mathbf{\Phi}}(x)$, the distribution of a component in the codebook $\mathbf{\Phi}$, as defined by (\ref{eqn:phiiid}).

Note that in the case of linear neurons and unitary recurrent matrix, the argument $K$ can be dropped, because there is no recency effect and all items in the sequence can be retrieved with the same accuracy. 

Thus, by inserting (\ref{eqn:z_mean}) and (\ref{eqn:z_var}) into (\ref{eqn:p_corr_full}), the accuracy for linear superposition becomes:
\begin{align}
\begin{split}
&p_{corr} = \int_{-\infty}^{\infty} \frac{dh}{\sqrt{2\pi}} e ^ {- \frac{1}{2} h^2} \times \\ 
&\left[ \Phi \left( \sqrt{\frac{M}{M-1+V_{\mathbf{\Phi}}(x^2)/V_{\mathbf{\Phi}}(x)^2}} \;h + \sqrt{\frac{N}{M-1+V_{\mathbf{\Phi}}(x^2)/V_{\mathbf{\Phi}}(x)^2}}  \right) \right] ^ {D-1}
\end{split}
\label{eqn:p_corr_lin_full}
\end{align}

Analogous to (\ref{eqn:p_corr_s}), for large $M$ the expression simplifies further to: 
\begin{equation}
p_{corr} = \int_{-\infty}^{\infty} \frac{dh}{\sqrt{2\pi}} e ^ {- \frac{1}{2} h^2} \left[ \Phi \left( h + s \right) \right] ^ {D-1} \; \mbox{ with } s=\sqrt{\frac{N}{M}}
\label{eqn:p_corr_lin_largeM}
\end{equation}
Interestingly, the expression (\ref{eqn:p_corr_lin_largeM}) is independent of the statistical moments of the coding vectors and thus applies to any distribution of coding vectors $p_{\mathbf{\Phi}}(x)$ (\ref{eqn:phiiid}).

\subsubsection{Analysis of VSA models from the literature}
\label{sec:vsa_literature}

Many connectionist models from the literature can be directly mapped onto the RNN and readout equations (\ref{eqn:rnn}, \ref{eqn:readout}). In the following, we will describe various VSA models and the properties of the corresponding encoding matrix $\mathbf{\Phi}$ and recurrent weight matrix $\mathbf{W}$. From this, we can derive for each of the frameworks the moments of $\mathbf{z}_{d,i}$ (\ref{eqn:z_mean}, \ref{eqn:z_var}), allowing us to compute retrieval accuracy $p_{corr}$ (\ref{eqn:p_corr_s}) and the total information $I_{total}$ (\ref{eqn:total_infoK}) in relation to the encoded sequence. 
 
In {\it hyperdimensional computing} (HDC) \citep{Gayler1998, Kanerva2009}, symbols are represented by $N$-dimensional random i.i.d. bipolar high-dimensional vectors (\emph{hypervectors}) and referencing is performed by a permutation operation. Thus the RNN (\ref{eqn:rnn}) corresponds to the HDC code of a sequence of $M$ symbols when the components of the encoding matrix $\mathbf{\Phi}$ are bipolar random i.i.d. $+1$ or $-1$, $p_\mathbf{\Phi}(x) \sim \mathcal{B} := \mathcal{U}\left( \lbrace -1, +1 \rbrace \right)$, and $\mathbf{W}$ is a permutation matrix, a special case of a unitary matrix.  

\begin{figure}[t]
\centering
\includegraphics[width=\textwidth]{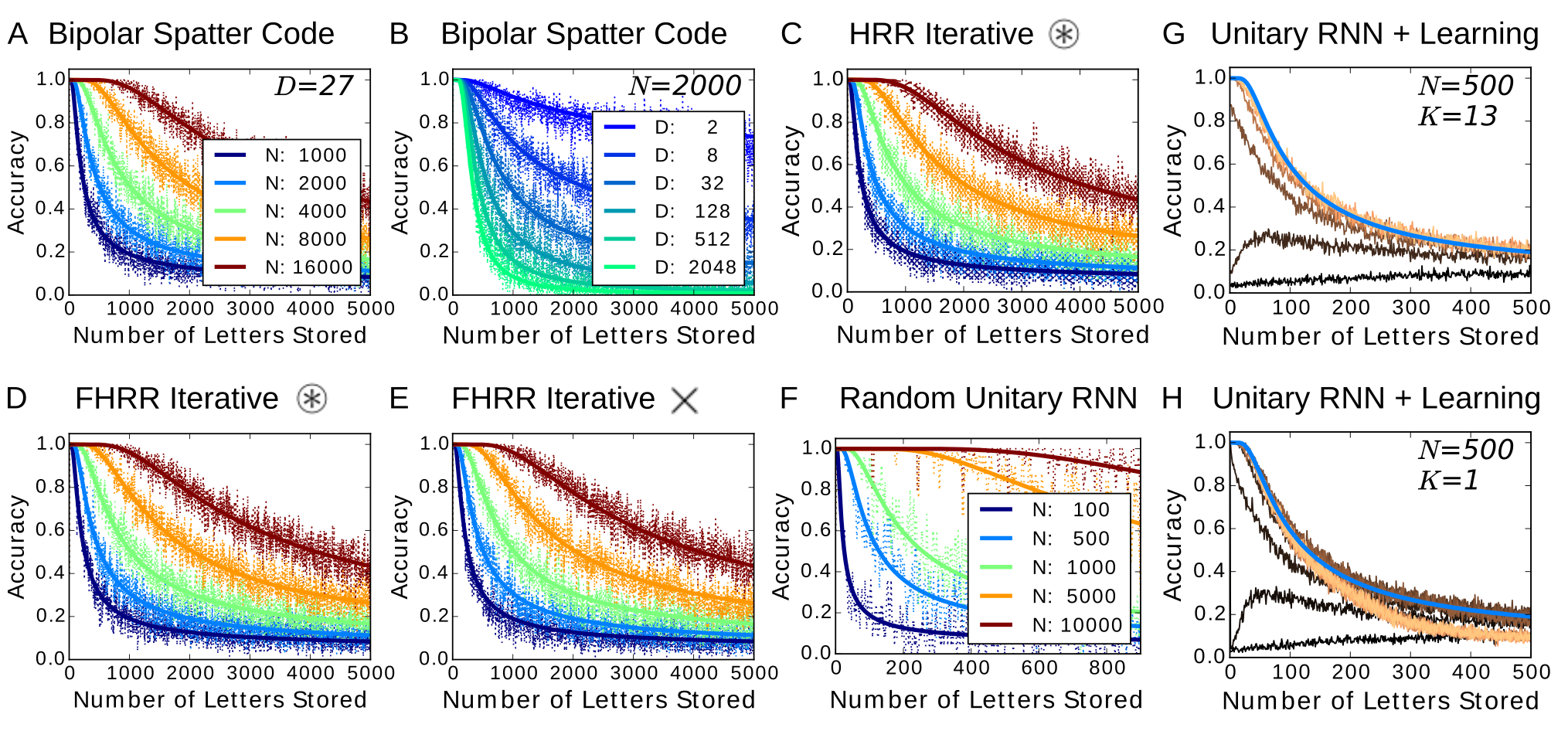}
\caption[Comparison between accuracy theory and simulation experiments]{\textbf{Classification retrieval accuracy: theory and simulation experiments.} The theory (solid lines) matches the simulation results (dashed lines) of the sequence recall task for a variety of VSA frameworks. 
Alphabet length in all panels except panel B is $D=27$.
A. Accuracy of HDC code as a function of the number of stored items for different dimensions $N$ of the hypervector. B. Accuracy of HDC with different $D$ and for constant $N=2000$. C. Accuracy of HRR code. D. Accuracy of FHRR code and circular convolution as the binding mechanism. E. Accuracy of FHRR using "multiply" as the binding mechanism. F. Accuracy achieved with random encoding and random unitary recurrent matrix also performs according to the same theory. G and H. Accuracy of RNN during training of the encoding matrix and decoding matrix with a fixed random unitary recurrent matrix. After 500 inputs the network is reset. As training time increases (black to copper), the performance converges to the theoretical prediction (blue line). With excessive training, the performance of the network can fall off in the low-fidelity regime (H).}
\label{fig:hd_capacity}
\end{figure}

With these settings, we can compute the moments of $\mathbf{z}_{d,i}$. We have $E_{\mathbf{\Phi}}(x^2) = 1$, $E_{\mathbf{\Phi}}(x)=0$,  $V_{\mathbf{\Phi}}(x^2)=0$ and $V_{\mathbf{\Phi}}(x)=1$, which can be inserted in equation (\ref{eqn:p_corr_lin_full}) to compute the retrieval accuracy. For large $M$ the retrieval accuracy can be computed using equation (\ref{eqn:p_corr_lin_largeM}).
We implemented this model and compared multiple simulation experiments to the theory.
The theory fits the simulations precisely for all parameter settings of $N$, $D$ and $M$ (Fig. \ref{fig:hd_capacity}A, B).

In {\it holographic reduced representation} (HRR) \citep{Plate1993, Plate2003}, symbols are represented by vectors drawn from a Gaussian distribution with variance $1/N$: $p_{\mathbf{\Phi}}(x) \sim \mathcal{N}(0, 1/N)$. The binding operation is performed by circular convolution and trajectory association can be implemented by binding each input symbol to successive convolutional powers of a random \emph{key} vector, $\mathbf{w}$. According to \citet{Plate1995}, the circular convolution operation can be transformed into an equivalent matrix multiply for a fixed vector by forming the circulant matrix from the vector (i.e. $ \mathbf{w} \circledast \mathbf{\Phi}_d = \mathbf{W} \mathbf{\Phi}_d $). This matrix has elements $\mathbf{W}_{ij}=\mathbf{w}_{(i-j) \% N}$ (where the subscripts on $\mathbf{w}$ a are interpreted modulo $N$). If $||\mathbf{w}|| = 1$, the corresponding matrix is unitary. Thus, HRR trajectory association can be implemented by an RNN with a recurrent circulant matrix and encoding matrix with entries drawn from a normal distribution. The analysis described for HDC carries over to HRR and the error probabilities can be computed through the statistics of $\mathbf{z}_{d,i}$, with $E_{\mathbf{\Phi}}(x)=0$, $E_{\mathbf{\Phi}}(x^2)=1/N$ giving $\mu_{d} = (1/N) \delta_{d=d'}$, and with $V_{\mathbf{\Phi}}(x)^2 = 1/N$, $V_{\mathbf{\Phi}}(x^2)=2/N$ giving $\sigma^2_{d} = (M+\delta_{d=d'})/N$. We compare simulations of HRR to the theoretical results in Fig. \ref{fig:hd_capacity}C.

The {\it Fourier holographic reduced representation} (FHRR) \citep{Plate2003} framework uses complex hypervectors as symbols, where each element is a complex number with unit radius and random phase, $p_{\mathbf{\Phi}}(x) \sim \mathcal{C} := \{ e^{i \mathcal{U}(0, 2\pi)} \}$. Technically, this has $N/2$ elements, but since each element is a complex number there are truly $N$ numbers in implementation (one for the real part and one for the imaginary part \citep{Danihelka2016}. This corresponds to the first $N/2$ positions of the encoding matrix $\mathbf{\Phi}$ acting as the real part, and the second $N/2$ acting as the imaginary part. Trajectory-association can be performed with a random vector with N/2 elements drawn from $\mathcal{C}$ acting as the key, raising the key to successive powers, and binding this with each input sequentially. 
In FHRR, both element-wise multiply or circular convolution can be used as the binding operation, and trajectory association can be performed to encode the letter sequence with either mechanism (see Methods \ref{sec:alt_vsa_analysis} for further details). These are equivalent to an RNN with the diagonal of $\mathbf{W}$ as the key vector or as $\mathbf{W}$ being the circulant matrix of the key vector.

Given that each draw from $\mathcal{C}$ is a unitary complex number $z=(cos(\phi), sin(\phi))$ with $p(\phi) \sim \mathcal{U}(0, 2\pi)$, the statistics of $\mathbf{z}_{d,i}$ are given by $E_{\mathbf{\Phi}}(x^2) = E(cos^2(\phi)) = 1/2$, $[E_{\mathbf{\Phi}}(x)]^2 = E(cos(\phi))^2 = 0$, giving $\mu_d=\delta_{d=d'}/2$. For the variance, let $ z_1 = (cos(\phi_1), sin(\phi_1))$ and $ z_2 = (cos(\phi_2), sin(\phi_2))$. Then $z_1^\top z_2 = \cos(\phi_1)\cos(\phi_2) + \sin(\phi_1)\sin(\phi_2) = cos(\phi_1 - \phi_2)$. Let $\phi_* = \phi_1 - \phi_2$, it is easy to see that it also the case that $p(\phi_*) \sim \mathcal{U}(0, 2\pi)$. Therefore, $V_{\mathbf{\Phi}}(x)^2=Var(cos(\phi_*))^2 = 1/4$ and $V_{\mathbf{\Phi}}(x^2)=0$ giving $\sigma^2_d = (M - \delta_{d=d'})/4$.
Again we simulate such networks and compare to the theoretical results (Fig. \ref{fig:hd_capacity}D, E). 

A random unitary matrix acting as a binding mechanism has also been proposed in the \textit{matrix binding with additive terms} framework (MBAT) \citep{Gallant2013}.
Our theory also applies to equivalent RNNs with random unitary recurrent matrices (created by QR decomposition of random Gaussian matrix), with the same $s=\sqrt{N/M}$ (Fig. \ref{fig:hd_capacity}F). 
Picking an encoding matrix $\mathbf{\Phi}$ and unitary recurrent matrix $\mathbf{W}$ at random satisfies the required assumptions (\ref{eqn:phiiid})-(\ref{eqn:wind}) with high probability when $N$ is large.

\subsubsection{RNN with learned encoding and decoding matrices}

To optimize accuracy, one can also use gradient descent for training the encoding and decoding matrices, $\mathbf{\Phi}$ and  $\mathbf{V}(K)$, in a linear RNN with a fixed random unitary recurrent matrix. The RNN with $N=1000$ is fed a sequence of random tokens and trained to recall the $K$th item in the sequence history, using the cross-entropy between the recalled distribution and the one-hot input distribution as the error function ($WTA$ in (\ref{eqn:readout}) is replaced with softmax). The network is evaluated each time step as more and more letters are encoded. After $500$ letters are presented, the network's activations are reset to 0 (see Methods (\ref{sec:rnn_train}) for further details). With training, the RNN can successfully recall the sequence history, but learning does not lead to any performance improvement beyond the theory for random distributed codes (Fig. \ref{fig:hd_capacity}G). Training the network for too long can even lead to performance loss in the low-fidelity regime (Fig. \ref{fig:hd_capacity}H).

\subsubsection{Detection accuracy of sparse input sequences}
\label{sec:detection}

\begin{figure}[t]
\centering
\includegraphics[width=\textwidth]{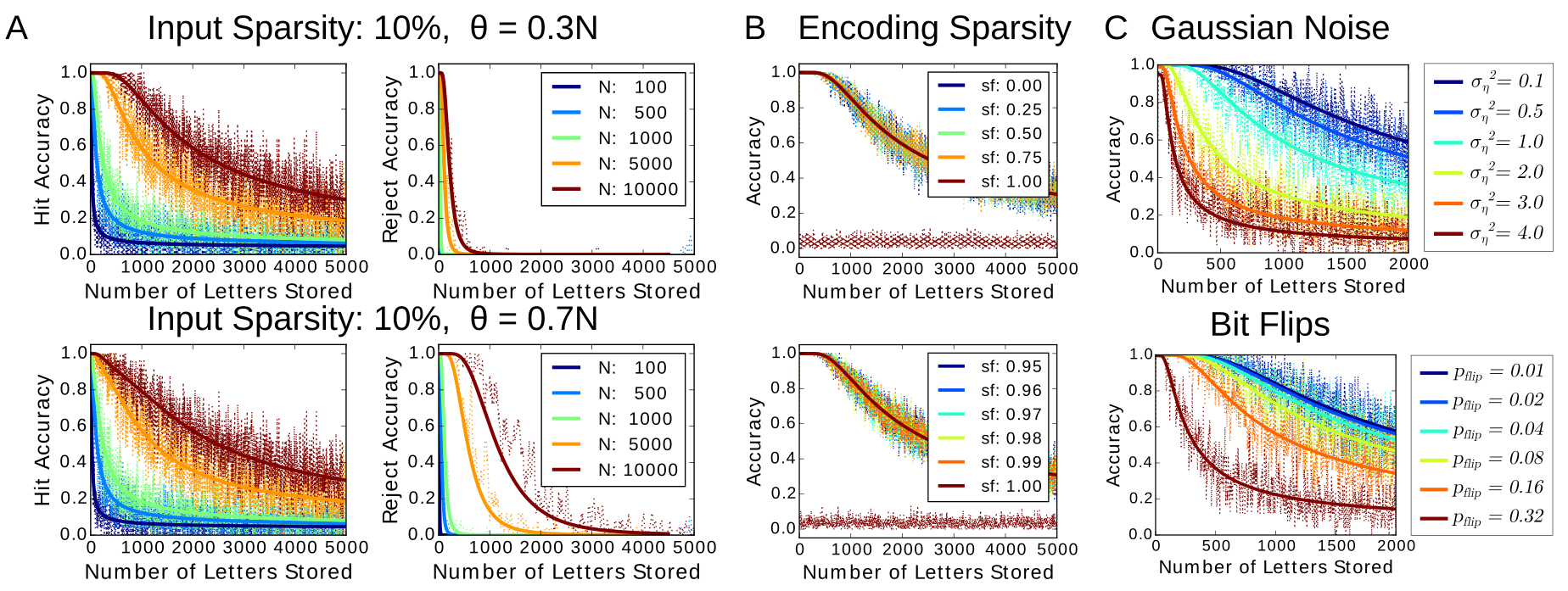}
\caption[Detection retrieval accuracy, encoding sparsity and noise]{\textbf{Detection retrieval accuracy, encoding sparsity and noise.} A. Retrieval of a sparse input sequence ($10\%$ chance for a zero vector). The hit and correct rejection performance for simulated networks (dashed lines) with different detection thresholds matches the theory (solid lines) -- a rejection is produced if $\mathbf{h}_d<\theta \ \forall d$.  B. Performance is not affected by the level of input sparsity until catastrophic failing when all elements are $0$. C. Simulations (dashed lines) match theory (solid lines) for networks corrupted by Gaussian noise (top) and random bit-flips (bottom). }
\label{fig:hd_capacity-extras}
\end{figure}

So far, an input is added in every time step of the RNN and the recall task is to select one of the $D$ tokens that was the correct input at a particular point in time. We next analyze the situation where only with some probability an input is added in a cycle. This probability is called \emph{input sparsity}. During detection, the network must first detect whether an input was present or not and, if present, determine its identity.  With a random encoding matrix, linear neurons and a unitary recurrent matrix, the SNR remains $s=\sqrt{N/M}$, but only when the input $\mathbf{a}$ generates a one-hot vector does this count towards incrementing the value of $M$.  The threshold setting trades off miss and false alarm error. We illustrate this in Fig. \ref{fig:hd_capacity-extras}A.

\subsubsection{Classification accuracy of sparsely encoded symbols}
\label{sec:encoding_sparsity}

Often neural activity is characterized as sparse and some VSA flavors utilize sparse codebooks, and several studies point to sparsity as an advantageous coding strategy for connectionist models \citep{Rachkovskij2001}. Sparsity in the encoding is considered by analyzing the same coding frameworks, but with many of the elements in the encoding matrix randomly set to $0$ with a probability called the ``sparseness factor'', $p_{sf}$. 
Sparsity affects both the signal and the noise equally, and essentially cancels out to lead to the same capacity result, $s=\sqrt{N/M}$. 
Sparsity essentially has no effect on the capacity of the hypervector, up until the point where there is a catastrophe when all of the entries are replaced by 0 (Fig. \ref{fig:hd_capacity-extras}B). 

\subsubsection{Classification accuracy in the presence of noise}
\label{sec:noise}

There are many ways to model noise in the network. Consider the case where there is only white noise added during the retrieval operation. We characterize this noise as i.i.d Gaussian with mean 0 and variance $\sigma_\eta^2$. It is easy to see that this noise will be added to the variance of $\mathbf{z}_{d,i}$, giving $s=\sqrt{N/(M+\sigma_\eta^2)}$. 
For the case where there is white noise added to the neural activations each time step, then the noise variance will accumulate with $M$, giving $s=\sqrt{N/(M(1+\sigma_\eta^2))}$.
If the noise was instead like a bit-flip, with the probability of bit-flip $p_{f}$, then this gives $s = \sqrt{\frac{N(1-2p_{f})^2}{M + 2p_{f}}}$. Finally, with these derivations of $s$ and (\ref{eqn:p_corr_s}), the empirical performance of simulated neural networks is matched (Fig. \ref{fig:hd_capacity-extras}C).

\subsubsection{Channel capacity of linear VSAs}
\label{sec:linear_scaling}

\begin{figure}[t]
\centering
\includegraphics[width=0.9\textwidth]{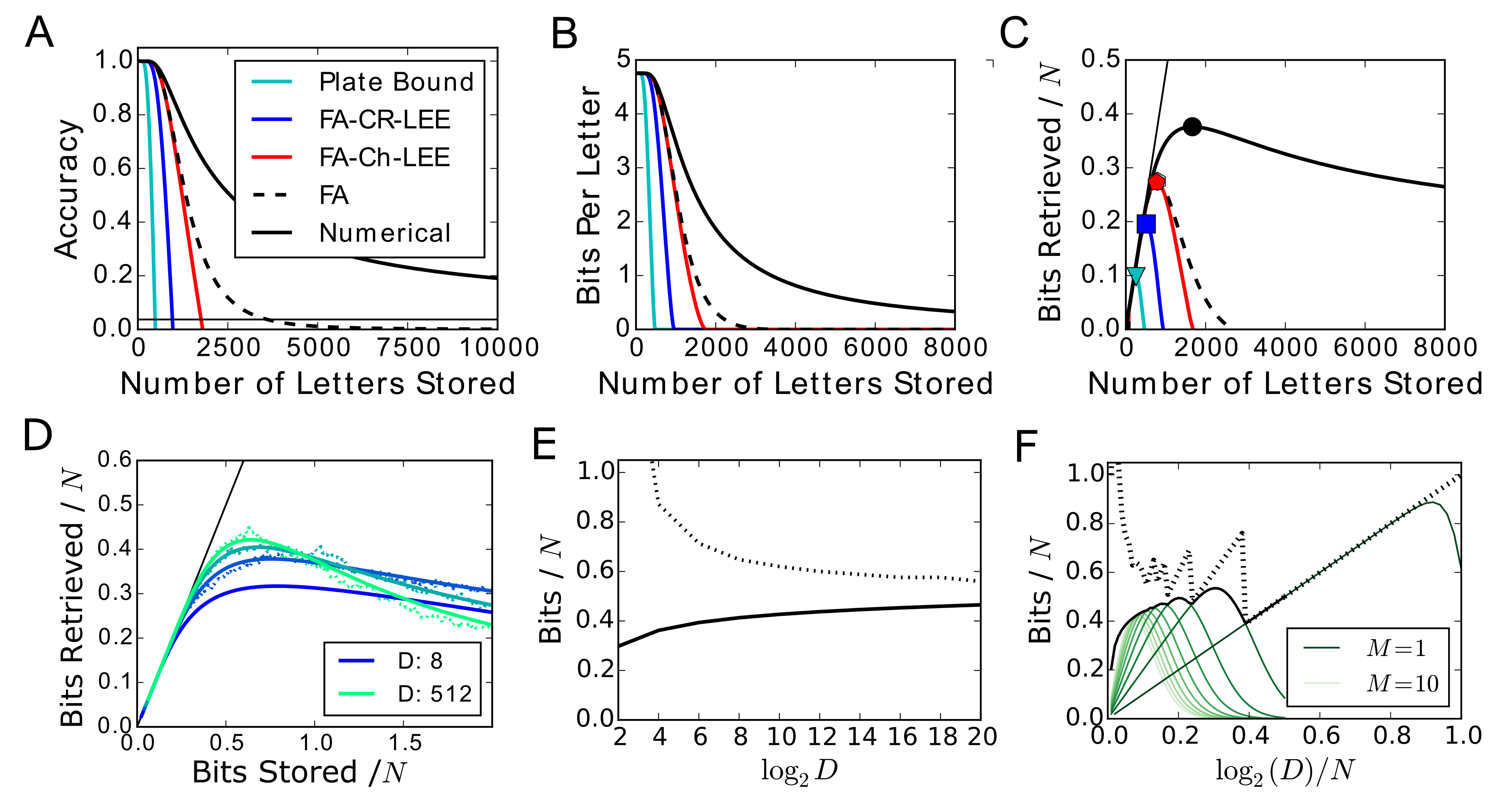}
\caption[Approximate linear relationships between dimensionality and capacity]{\textbf{Information storage and channel capacity.} A. Approximations of retrieval accuracy derived in Results \ref{sec:p_corr_approx} and \citet{Plate1993} are compared to the numerically evaluated accuracy ($p_{corr}$). The approximations all underestimate the accuracy in the low fidelity regime. B. The information per item (\ref{eqn:info_item}). C. The total information that can be retrieved, and channel capacity (solid points) predicted by different theories. D. Retrieved information measured in simulations (dashed lines) compared to the predictions of the theory (solid lines). The information maximum is dependent on $D$, and is linear with $N$. E. Information retrieval maximum for different values of $D$ (solid line) and bits stored at retrieval maximum ($I_{stored}$ computed with (\ref{eqn:total_infoK}) -- dashed line). F. Maximum information retrieved (solid black line) and total information stored ($I_{stored}$ -- dashed black) in the range where $D$ is a significant fraction of $2^N$ ($N=100$). In addition, the retrieved information for fixed values $M=\{1, ..., 10 \}$ is plotted (green lines of different shades). The different local maxima in the black line correspond to small fixed values of $M$: For $M=1$, retrieved and stored information come close to each other near 1 bit per neuron. With larger $M$, the gap between the two curves grows. 
  }
\label{fig:linear_fit-info}
\end{figure}

The original estimate for the capacity of distributed random codes \citep{Plate1993} was based on the same kind of approximation to $p_{corr}$ in Results \ref{sec:p_corr_approx}, but considered a different setup (Methods \ref{sec:plate_compare}). The \citet{Plate1993} approximation only differs from the FA-CR-LEE approximation (\ref{eqn:nm_chernoff}) by a factor of 2, due to efforts to produce a lower bound. Figure \ref{fig:linear_fit-info}A compares the approximations (\ref{eqn:factorialapro}, \ref{eqn:p_corr_apx}, \ref{eqn:nm_chernoff}) with the true numerically evaluated integral (\ref{eqn:p_corr_s}). These approximations are good in the high-fidelity regime, where $p_{corr}$ is near 1, but underestimate the performance in the low-fidelity regime. 
 
With the relationship $s=\sqrt{N/M}$, the information contained in the activity state $\mathbf{x}(M)$ can be calculated. We compare the information per item stored (Fig. \ref{fig:linear_fit-info}B; \ref{eqn:info_item}) and the total information (\ref{eqn:total_infoK}) based on the approximations with the true information content determined by numeric evaluation of $p_{corr}$. 
In the linear scenario with unitary weight matrix, $p_{corr}$ has no dependence on $K$, and so the total information in this case is simply $I_{total} = M I_{item}$ (\ref{eqn:total_infoK}).

The channel capacity based on \citet{Plate2003}'s approximation reaches about 0.11 bits per neuron (Fig. \ref{fig:linear_fit-info}C; \citet{Plate2003}'s approximation is better for higher $D$, and reaches as high as 0.15 bits per neuron; Methods \ref{sec:plate_compare}). The new high-fidelity approximations bring us as high as 0.27 bits per neuron with $D=27$, but all underestimate the low-fidelity regime where the true information maximum of the network is located. We numerically evaluated $p_{corr}$ and see that the true channel capacity of the network is nearly 0.4 bits per neuron with $D=27$ (Fig. \ref{fig:linear_fit-info}C, black circle). The true information maximum is four times larger than the previous approximation would suggest, but exists well into the low-fidelity regime

The theory precisely fits simulations of the random sequence recall task with different $D$ and empirically measured total information (Fig. \ref{fig:linear_fit-info}D). The total information per neuron scales linearly with the number of neurons, and the maximum amount of information per element that can be stored in the network is dependent upon $D$. The channel capacity increases further as $D$ becomes larger, to over 0.5 bits per neuron with large $D$ (Fig. \ref{fig:linear_fit-info}E, solid line). 

As the total number of unique tokens, $D$, grows larger (up to $2^N$) $M$ gets smaller to maximize the information content (Fig. \ref{fig:linear_fit-info}E, dashed line). The theory eventually breaks down when there is no superposition, i.e. when $M=1$. 
The hit distribution, $\mathbf{h}_{d'}$ has no variance when only one item is stored in the network, with fixed value $\mathbf{h}_{d'} = N E_{\mathbf{\Phi}}[x^2]$. Further, it is not acceptable to assume that the other $\mathbf{h}_d$ values are random Gaussian variables, because $\mathbf{h}_{d'}$ is actually the maximum possible of any $\mathbf{h}_d$ and there is 0 probability that a greater value will be drawn. Thus, when $M=1$ there is only confusion when two of the $D$ codewords are exactly the same, or there is a \emph{collision}. 

If the codewords are independently drawn with $p_{\mathbf{\Phi}} \sim \mathcal{U}(\lbrace -1, +1\rbrace)$, and collision ties are decided uniformly random, then the probability of accurately identifying the correct codeword is:
\begin{equation}
p_{corr}^{M=1} = \sum_c p_c / (c + 1)
\label{eqn:p_corr_M1}
\end{equation}
where $p_c$ is the probability of a vector having collisions with $c$ other vectors in the codebook of $D$ vectors, which can be found based on the binomial distribution:
\begin{equation}
p_c = \binom{D}{c} q^c (1-q)^{D-c}
\end{equation}
where $q=1/2^N$ is the likelihood of a pair of vectors colliding. Without superposition, that is, for $M=1$, the collisions reduce the accuracy $p_{corr}^{M=1}$ to $(1 - 1/e) \approx 0.63$ for $D = 2^N$ in the limit $N \to \infty$. For examples of this reduction at finite size $N$, see  Fig. \ref{fig:linear_fit-info}F and Methods Fig. \ref{fig:largeD}A. In the presence of superposition, that is, for $M>1$, the cross talk noise becomes the limiting factor for channel capacity. The resulting channel capacity exceeds $0.5$ bits per neuron for small $M$ and is still above $0.2$ bits per neuron for larger $M$-values (\ref{fig:linear_fit-info}F, black line). The capacity curves for fixed values of $M$ (\ref{fig:linear_fit-info}F, green lines) show the effect of cross talk noise, which increases as more items are superposed (as $M$ increases). For $M=1$ (\ref{fig:linear_fit-info}F, dark green line), equations (\ref{eqn:p_corr_M1}) and (\ref{eqn:total_infoK}) can be evaluated as $D$ grows to $2^N$. The theory derived here is specifically describing random-distributed superposition, and we elaborate on coding without superposition where $M=1$ further in Methods \ref{sec:M1}. 

\subsection{Networks exhibiting recency effects}
\label{sec:recency}

With the linear networks described in the previous section, there is no \emph{recency effect}, which means that the readout of the most recent letter stored is just as accurate as the earliest letter stored. However, networks with non-linear activation function or a recurrent weight matrix that is not unitary will have a recency effect. Through the recency effect, sequence items presented further back in time will be forgotten. Forgetting can be beneficial because new inputs can be continuously stored without relying on additional reset mechanisms. In this section, the theory extends to RNNs that exhibit recency effects and we generalize the mapping between RNNs and VSAs.  

\subsubsection{Linear neurons and contracting recurrent weights}
\label{sec:eigen_decay}
 
Consider a network of linear neurons with recurrent weight matrix $\mathbf{W} = \lambda \mathbf{U}$, $\mathbf{U}$ unitary, in which the attenuation factor $0<\lambda<1$ contracts the network activity in each time step. After a sequence of $M$ letters has been applied, the variance of $\mathbf{z}_{d,i}$ is $\left(\frac{1-\lambda^{2M}}{1-\lambda^2} \right) V_{\mathbf{\Phi}}(x)^2$, and the signal decays exponentially with $\lambda^{K} E_{\mathbf{\Phi}}(x^2)$. The SNR for recalling the input that was added $K$ time steps ago is:
\begin{equation}
s (K) = \lambda^K \sqrt{\frac{N (1 - \lambda^2)}{1-\lambda^{2M}}}
\label{eqn:M_snr}
\end{equation}

\begin{figure}[p]
 \centering
\includegraphics[width=\textwidth]{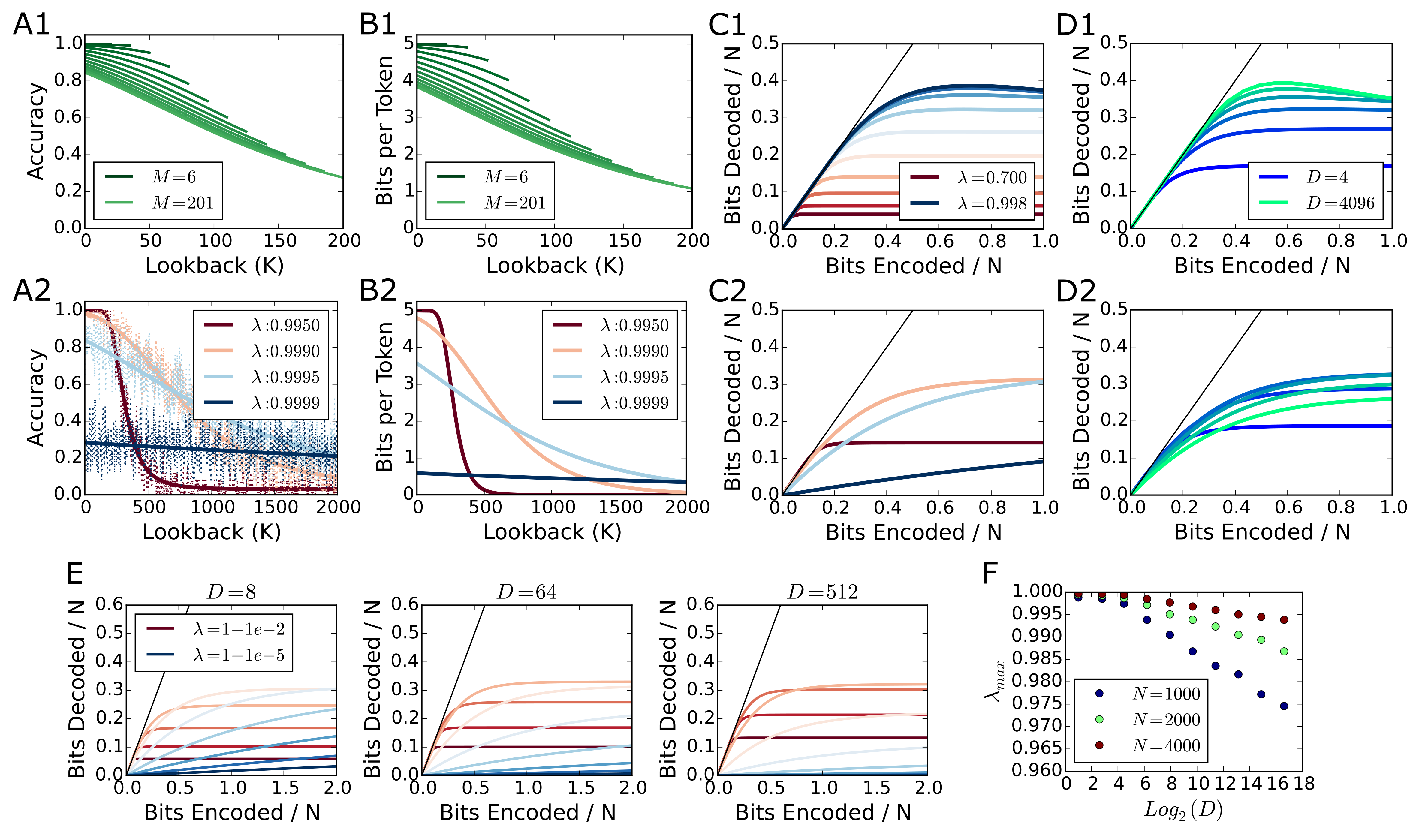}
\caption[Information decay]{\textbf{Linear network with contracting recurrent weights.} A1. Accuracy in networks with $\lambda < 1$. Multiple evaluations of $p_{corr}$ are shown as a function of $K$ for sequences of different lengths, $M$.  ($\lambda=0.996$, $N=1000$). B1. The information per item $I_{item}$ also depends on $K$. C1. The total retrieved information per neuron for different $\lambda$. The maximum is reached as $\lambda$ approaches $1$ when $M$ is finite ($D=64$; $N=1000$). D1. The retrieved information is maximized as $D$ grows large ($\lambda=0.988$). A2. Accuracy in networks with $\lambda<1$ as $M \to \infty$ ($N=10,000$; $D=32$). B2. Information per item. C2. Total information retrieved as a function of the total information stored for different $\lambda$. There is a $\lambda$ that maximizes the information content for a given $N$ and $D$ ($D=64$). D2.
Total information retrieved as a function of the total information stored for different $D$ ($\lambda=0.999$). Retrieved information is optimized by a particular combination of $D$ and $\lambda$. E. The total retrieved information per neuron versus the information stored per neuron for different $D$ and $\lambda$ over a wide range. As $D$ increases the information is maximized by decreasing $\lambda$. F. Numerically determined $\lambda_{max}$ values that maximize the information content of the network with $M \to \infty$ for different $N$ and $D$. }
\label{fig:decay_info}
\end{figure}

Thus, the SNR decays exponentially as $K$ increases, and the highest retrieval SNR is from the most recent item stored in memory. The accuracy (Fig. \ref{fig:decay_info}A1) and information per item (Fig. \ref{fig:decay_info}B1) based on this formula for $s(K)$ shows the interdependence between the total sequence length ($M$) and the lookback distance ($K$) in the history. 

Equation (\ref{eqn:M_snr}) is monotonically increasing as $\lambda$ increases, and thus to maximize the SNR for the $K$-th element in history given a finite set of $M$ stored tokens, we would want to maximize $\lambda$, or have the weight matrix remain unitary with $\lambda=1$ \footnote{If we allowed $\lambda$ to be larger than $1$, then memories from the past would grow in magnitude exponentially -- this would mean higher SNR for more distant memories at the cost of lower SNR for recent memories (this would cause the network to explode, however normalization could be used.)}. The channel capacity is maximized as $\lambda \to 1$ when $M$ is finite (Fig. \ref{fig:decay_info}C1) and as $D$ grows large (Fig. \ref{fig:decay_info}D1).

With eigenvalues less than one, the memory can operate even when it is exposed to a continuous stream of inputs without indication when a computation ends. Thus, $M$ grows infinitely large, $M \to \infty$, but also the signals of the past decay away exponentially. Depending on the value of $\lambda$, crosstalk noise and signal decay influence the SNR. For large $M$, the noise variance is bounded by $\frac{1}{1-\lambda^2}$, and the network reaches its \emph{filled} equilibrium state. The SNR for the $K$-th element back in time from 
the filled state is:
\begin{equation}
s(K) = \lambda^K \sqrt{N (1 - \lambda^2)}
\label{eqn:eigen_snr}
\end{equation}

Simulated networks that store a sequence of tokens via trajectory association with contracting recurrent weights and where $M >> N$ (Fig. \ref{fig:decay_info}A2, solid lines) match the theory from (\ref{eqn:eigen_snr}) and (\ref{eqn:p_corr_s}) for different $\lambda$ (Fig. \ref{fig:decay_info}A2, dashed lines). The information per item retrieved (Fig. \ref{fig:decay_info}B2) and the total information (Fig. \ref{fig:decay_info}C2) for different values of $\lambda$ shows the trade-off between fidelity and duration of storage, and that there is an ideal decay value that maximizes the channel capacity of the memory buffer with $M \to \infty$ for a given $N$ and $D$. This ideal decay value differs depending on the number of potential tokens ($D$; Fig. \ref{fig:decay_info}D2). When more potential tokens are present (meaning more bits per item), then the activity should be decayed away more quickly and a have a shorter history to optimize information content (Fig. \ref{fig:decay_info}E). The decay eigenvalues for different $N$ and $D$ that maximize the channel capacity were computed numerically and reveal a relationship to other network parameters (Fig. \ref{fig:decay_info}F).

\subsubsection{Neurons with clipped-linear transfer function}
\label{sec:non_linear}

Non-linear activation functions, $f(\mathbf{x})$ in (\ref{eqn:rnn}), also induce a recency effect. Consider the clipped-linear activation function, $f_\kappa$, in which clipping prevents the neurons from exceeding $\kappa$ in absolute value:
\begin{equation}
\mathbf{x} (m) = f_\kappa \left(\mathbf{W} \mathbf{x} (m-1) + \mathbf{\Phi} \mathbf{a} (m) \right)
\label{eqn:clip_upd}
\end{equation}
where:
\begin{equation}
f_\kappa (x) = 
\begin{cases}
-\kappa & x \leq -\kappa \\
x & -\kappa < x < \kappa \\
\kappa & x \geq \kappa
\end{cases}
\end{equation}

Such a non-linear function plays a role in VSAs which constrain the activation of memory vectors, such as the \emph{binary-spatter code} \citep{Kanerva1996} or the \emph{binary sparse-distributed code} \citep{Rachkovskij2001}, but we consider the clipping function more generally when mapped to the RNN.

With standard HDC encoding, using bipolar random codes and permutation matrix, the components of $\mathbf{x}$ will always assume integer values. Due to the clipping, the activations of $\mathbf{x}$ are now confined to $\{-\kappa, -\kappa +1,..., \kappa\}$. As a consequence, $\mathbf{z}$ will also assume values limited to $\{-\kappa, ..., \kappa \}$.  To compute $s$, we need to track the mean and variance of $\mathbf{z}_{d,i}$. To do so, we introduce a vector notation $\mathbf{p}_{\mathcal{J}(k)}(m) := p(\mathbf{z}_{d,i}(m) = k) \ \forall k \in \{ -\kappa, ..., \kappa \}$, which tracks the probability distribution of $\mathbf{z}_{d,i}$. The probability of each of the integers from $\{-\kappa, ..., \kappa\}$ is enumerated in the $2\kappa+1$ indices of the vector $\mathbf{p}$, and $\mathcal{J}(k) = k + \kappa$ is a bijective map from the values of $\mathbf{z}$ to the indices of $\mathbf{p}$ with inverse $\mathcal{K}=\mathcal{J}^{-1}$. To compute the SNR of a particular recall, we need to track the distribution of $\mathbf{z}_{d,i}$ with $\mathbf{p}$ before the item of interest is added, when the item of interest is added, and in the time steps after storing the item of interest.

{\it Storage after resetting:} At initialization $\mathbf{x}_i(0) = 0 \ \forall i$, and so $\mathbf{p}_{j}(0) = \delta_{\mathcal{K}(j)=0}$. For each time step that an input arrives in the sequence prior to the letter of interest, a $+1$ or $-1$  will randomly add to $\mathbf{z}_{d,i}$ up until the bounds induced by $f_\kappa$, and this can the be tracked with the following diffusion of $\mathbf{p}$:
\begin{equation}
 \mathbf{p}_{j} \left( m+1 \right)
 =  \frac{1}{2} 
\begin{cases} 
 \mathbf{p}_{j}(m)+\mathbf{p}_{j+1}(m)    &  \mbox{  when }  \mathcal{K}(j)=-\kappa  \\
 \mathbf{p}_{j-1}(m)+\mathbf{p}_{j} (m)   &  \mbox{  when } \mathcal{K}(j)=\kappa \\
 \mathbf{p}_{j-1}(m) +\mathbf{p}_{j+1} (m)  &  \mbox{  otherwise. }
\end{cases}
\ \forall \ m \ne M-K
\label{eqn:pj_upd} 
\end{equation}

Once the vector of interest arrives at $m=M-K$, then all entries in $\mathbf{z}_{d,i}$ will have $+1$ added.
This causes the probability distribution to skew:
\begin{equation}
 \mathbf{p}'_{j} \left( m+1 \right)
 = 
\begin{cases} 
 0    &  \mbox{  when }  \mathcal{K}(j)=-\kappa  \\
 \mathbf{p}_{j}(m)+\mathbf{p}_{j-1} (m)   &  \mbox{  when } \mathcal{K}(j)=\kappa \\
 \mathbf{p}_{j-1}(m)  &  \mbox{  otherwise. }
\end{cases}
\ m = M-K
\label{eqn:pj_skew}
\end{equation}

The $K-1$ letters following the letter of interest, will then again cause the probability distribution to diffuse further based on (\ref{eqn:pj_upd}). Finally, $s$ can be computed for this readout operation by calculating the mean and variance with $\mathbf{p} (M)$:
\begin{equation}
\mu_d = \delta_{d=d'} \sum_{j=0}^{2 \kappa} \mathcal{K}(j) \mathbf{p}_{j} (M) 
\label{eqn:nl_mean}
\end{equation}
\begin{equation}
\sigma^2_d = \sum_{j=0}^{2 \kappa} (\mathcal{K}(j) - \mu_d)^2 \mathbf{p}_{j}(M)
\label{eqn:nl_var}
\end{equation}

{\it Storage without resetting:} For $M \to \infty$ the diffusion equation (\ref{eqn:pj_upd}) will reach a flat equilibrium distribution, with the values of $\mathbf{z}_{d,i}$ uniformly distributed between $\{-\kappa, ..., \kappa \}$: $\mathbf{p}_{j}(\infty) = 1 / (2\kappa + 1) \ \forall j$. This means, like with eigenvalue decay, the clipped-linear function bounds the noise variance. Thus, information can still be stored in the network even after being exposed to an infinite sequence of inputs. In this case, the variance of $\mathbf{z}_{d,i}$ reaches its maximum, the variance of the uniform distribution, $((2\kappa + 1)^2 -1)/12$. Accordingly, the SNR can be calculated with $M\to \infty$ by replacing in (\ref{eqn:pj_skew}) $\mathbf{p}(m)$ with $\mathbf{p}(\infty)$, and then again using the diffusion equation (\ref{eqn:pj_upd}) for the $K-1$ items following the item of interest.

\begin{figure}[tp]
\centering
\includegraphics[width=\textwidth]{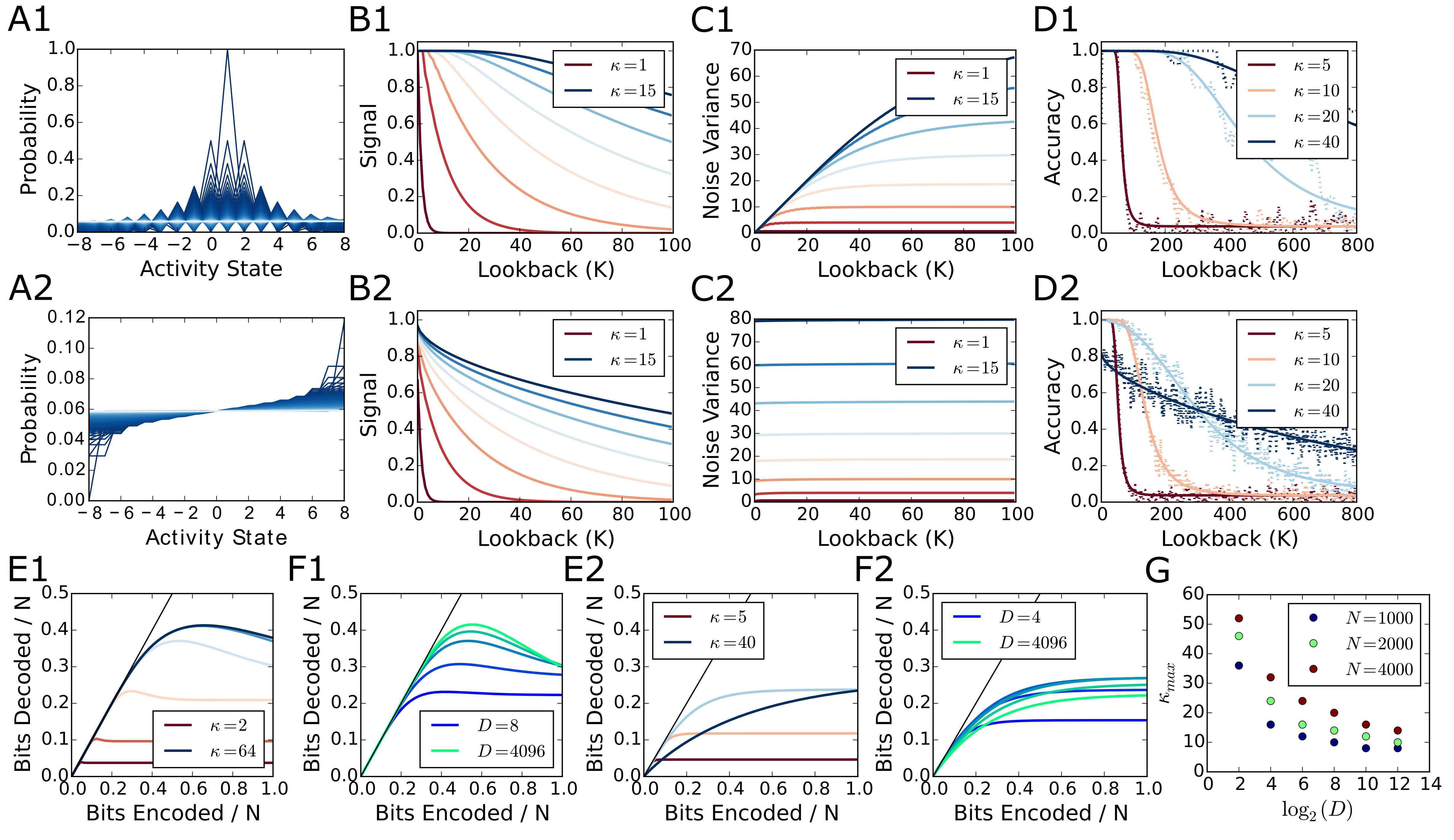}
\caption[Clipping capacity]{\textbf{Capacity for neurons with clipped-linear transfer function.} A1. The probability distribution of one term $\mathbf{z}_{d,i}$ in the inner product used for retrieval of the first sequence item, as the sequence length is increased. The distribution evolves according to (\ref{eqn:pj_upd}) and (\ref{eqn:pj_skew}), it begins at a delta function (dark blue), and approaches the uniform equilibrium distribution when $M$ is large (light blue). B1. The clipped-linear function causes the signal to degrade as more items are stored in the network ($M=K$; $N=5000$; $D=27$). C1. The variance of the distribution grows as more items are stored, but is bounded by the clipped-linear function. D1. The accuracy theory fits empirical simulations decoding the first input as more letters are stored (dashed lines; $M=K$; $N=5000$; $D=27$). A2. The probability distribution of $\mathbf{z_{d,i}}$ when a new item is entered at full equilibrium, that is, when $M \to \infty$. The most recent letter encoded (dark blue) has the highest skew, and the distribution is more similar to the uniform equilibrium for letters further in the past of the sequence (light blue). B2. The signal is degraded from crosstalk, and decays as a function of the time gap between encoding and recall (lookback). C2. The noise variance is already saturated and stays nearly constant as a function of the time gap. D2. The accuracy exhibits a trade-off between fidelity and memory duration governed by $\kappa$. E1. When $M$ is finite, the information that can be decoded from the network reaches a maximum when $\kappa$ is large ($D=256$). F1. The capacity increases with $D$ ($\kappa=20$). E2. When $M \to \infty$, there is a trade-off between fidelity and memory duration, a particular $\kappa$ value maximizes the retrieved information for a given $D$ and $N$ ($D=256$). F2. For a given memory duration ($\kappa=20$) an intermediate $D$ value maximizes the retrieved information. G. The memory duration $\kappa_{max}$ that maximizes the retrieved information. 
}
\label{fig:clipping_capacity}
\end{figure}

Figure \ref{fig:clipping_capacity} illustrates this algorithm for tracking the distribution of $\mathbf{z}_{d,i}$ activity states. When the item of interest is added, the probability distribution is most skewed and the signal degredation is relatively small. As more items are added later, the distribution diffuses to the uniform equilibrium, and the signal decays to 0. The figure shows the network operating at the two extrema: empty (Fig. \ref{fig:clipping_capacity} A1-F1) and filled (Fig. \ref{fig:clipping_capacity} A2-F2). Found numerically for different $N$ and $D$, the parameter $\kappa_{max}$ maximizes the information content of the network when $M \to \infty$ (Fig. \ref{fig:clipping_capacity} G).

\subsubsection{Neurons with sqashing non-linear tranfer functions}
\label{sec:general_nl}

The case when the neural transfer function $f(\mathbf{x})$ is a saturating or sqashing function with $|f(x)|$ bounded by a constant also implies $\mathbf{z}_{d,i}$ is bounded and $|\mathbf{z}_{d,i}| \leq \kappa$. For any finite fixed error, one can choose an $n$ large enough so that the distribution $p(\mathbf{z}_{d,i}=k) = \mathbf{p}_{\mathcal{J}(k)}$ can be approximated by discretizing the state space into $2n+1$ equal bins in $\mathbf{p}$. Similar as for the clipped-linear transfer function, one can construct a bijecteve map from values to indices and track $\mathbf{p}$ approximately using rounding to discretize the distribution, $\mathcal{J}(k)= \lfloor \frac{n}{\kappa} (k+\kappa) \rceil$, with inverse $\mathcal{K}=\mathcal{J}^{-1}$. The transition kernel update (\ref{eqn:det_sys}) can be simplified to two updates, one for the signal and one for the distractors:
\begin{equation}
    \mathbf{p}_{j^*}(m+1) = \sum_{j=0}^{2n} \int dy \ p_{\mathbf{\Phi}}(y) \mathbf{p}_{j} (m) \ \delta_{j^*=\mathcal{J}(f(\mathcal{K}(j)+y))}
\ \ \ \forall m \ne M-K
\label{eqn:p_distractor_av}
\end{equation}

The update for the signal, given at $m=M-K$, where we know that $\mathbf{\Phi}_{d'}$ was stored in the network:
\begin{equation}
    \mathbf{p}'_{j^*}(m+1) = \sum_{j=0}^{2n} \int dy \ p_{\mathbf{\Phi}}(y) \mathbf{p}_{j} (m) \ \delta_{j^*=\mathcal{J}(f(\mathcal{K}(j)+y^2))}
\ \ \ m = M-K
\label{eqn:p_signal_av}
\end{equation}

We illustrate our analysis for $f(x) = \gamma\tanh(x/\gamma)$, where $\gamma$ is a free gain parameter, and for the HDC code with permutation matrix. With these choices, the network update for each time step is:
\begin{equation}
\mathbf{x} (m) = \gamma \tanh \left( \frac{\mathbf{W} \mathbf{x}(m-1) + \mathbf{\Phi} \mathbf{a} (m) }{\gamma} \right)
\label{eqn:tanh_upd}
\end{equation}

\begin{figure}[t]
\centering
\includegraphics[width=\textwidth]{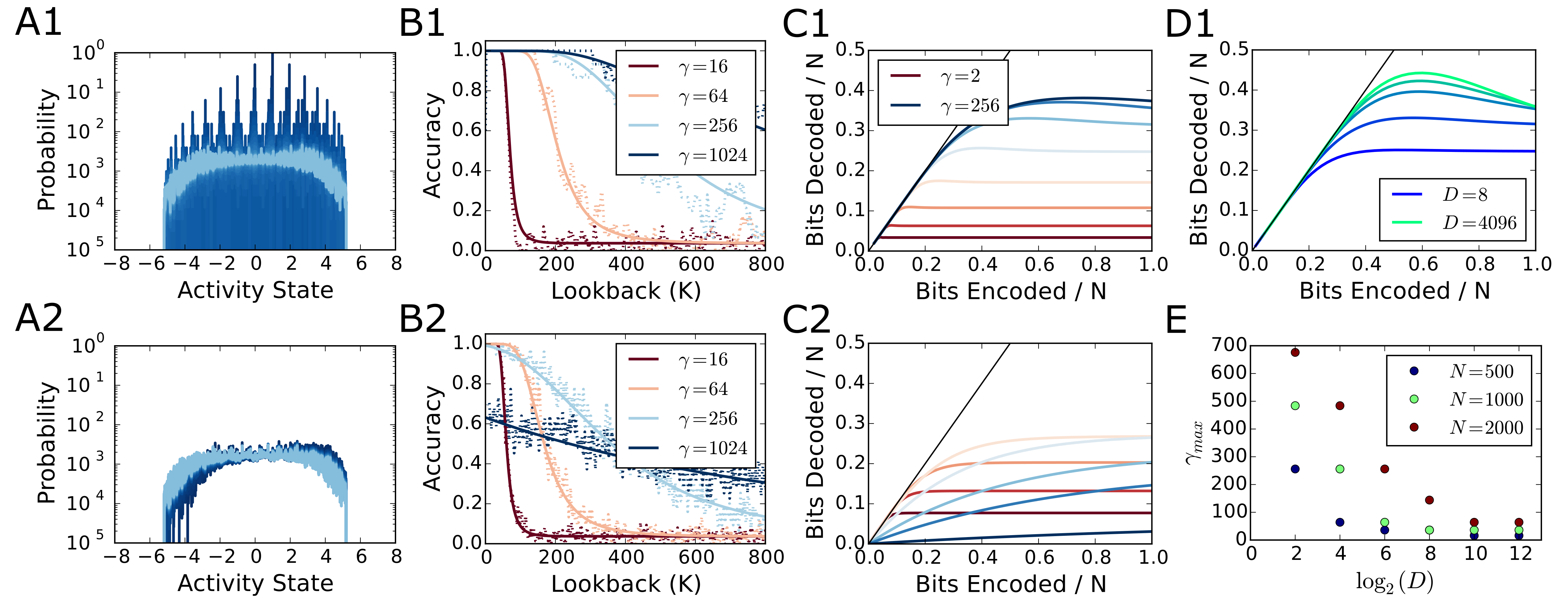}
\caption[Clipping capacity]{\textbf{Capacity for neurons with saturating non-linearity} A1. The probability distribution of one term $\mathbf{z}_{d,i}$ in the inner product used for retrieval of the first sequence item, as the sequence length is increased. The distribution begins as a delta function (dark blue) when only one letter is stored, and approaches the equilibrium distribution when $M$ is large (light blue). B1. The accuracy theory (solid lines) correctly describes simulations (dashed lines) retrieving the first input as more letters are stored ($M=K$; $N=2000$; $D=32$). C1. Retrieved information as a function of the stored information. When $M$ is finite, the maximum is reached for large $\gamma$ ($D=256$). D1. The capacity increases as $D$ increases ($\gamma=64$). A2. The probability distribution of $\mathbf{z_{d,i}}$ when a new item is entered at full equilibrium, that is, when $M \to \infty$. The distribution for most recent letter posesses the highest skew (dark blue), and the distribution is closer to the uniform equilibrium (light blue) for letters encoded further back in the history. B2. The accuracy exhibits a trade-off between fidelity and memory duration governed by $\gamma$. C2. When $M$ is large, there is a $\gamma$ that maximizes the information content for a given $D$ and $N$ ($D=256$). E. Numerically computed $\gamma_{max}$ that maximizes the information content.}
\label{fig:tanh_capacity}
\end{figure}

The simulation experiments with the $\tanh$ activation function examined both initialization extrema, empty (Fig. \ref{fig:tanh_capacity} Row 1) and filled (Fig. \ref{fig:tanh_capacity} Row 2). The iterative approximation algorithm to compute $s$ shows $\tanh$ has similar effects as the clipping activation function. 

\subsubsection{Memory buffer time constant and optimization}

Contracting weights and non-linear activation functions both induce a recency effect by decaying the signal and bounding the noise variance. The \emph{buffer time constant} ($\tau$) can be derived from the exponential decay of contracting recurrent weights based on $\lambda^K=e^{-K/\tau}$ (\ref{eqn:eigen_snr}):
\begin{equation}
\tau (\lambda) = -1 / \log \lambda
\label{eqn:tau_lambda}
\end{equation}

\begin{figure}
\centering
\includegraphics[width=\textwidth]{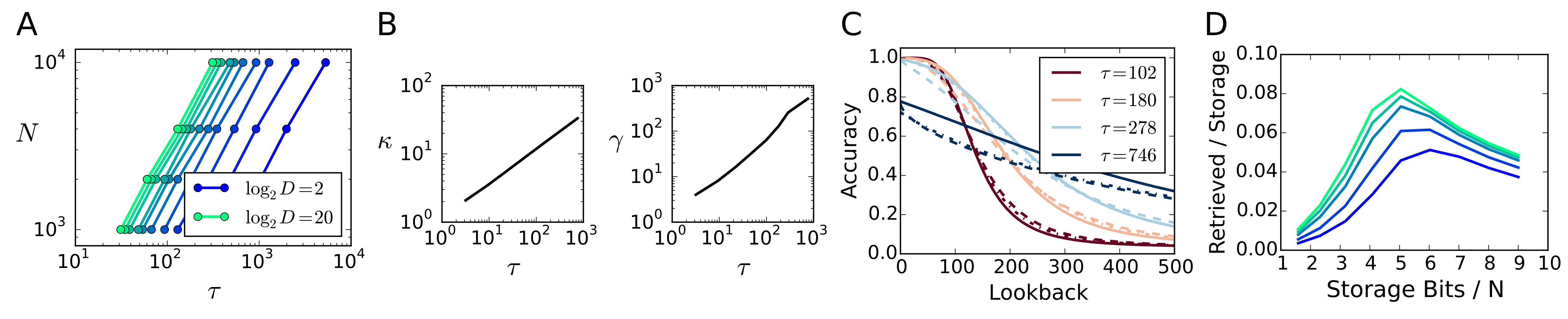}
\caption[Clipping information analysis]{\textbf{Buffer time constant and optimization.} A. The optimal $N$ for given time constant $\tau$ and $D$. B. The relationship between time constant and non-linear activation parameters, $\kappa$ and $\gamma$. C. Accuracy comparison of decay (solid), clipped-linear (dashed) and $\tanh$ (dotted) networks which share the same bound of noise variance. D. Ratio between retrieved and stored information for the clipped HDC network. The ratio is optimized with 4-5 bits of resolution per element ($D=[8, 32, 256, 1024, 4096]$, dark to light). }
\label{fig:tau_info}
\end{figure}

We derive the $N$ that optimizes the amount of information in the memory buffer for a desired time constant (Fig. \ref{fig:tau_info}A).
The time constant for non-linear activation functions can also be approximated by equating the bound of the variance induced by the non-linearity to the bound induced by $\lambda$. For clipping, the noise variance is bounded by the variance of the uniform distribution, $((2\kappa + 1)^2-1)/12$, which can be equated to the bound of $1/(1-\lambda^2)$ and with (\ref{eqn:tau_lambda}) gives:
\begin{equation}
\tau (\kappa) = \frac{-2}{\log\left(1-\frac{3}{\kappa(\kappa+1)}\right)}
\end{equation}

Relating the bound of the noise variance to $1/(1-\lambda^2)$ is a general technique to approximate the time constant for any non-linear function with (\ref{eqn:tau_lambda}). We show the relationship between $\tau$ with both the clipped-linear parameter $\kappa$ (\ref{eqn:clip_upd}), as well as the $\tanh$ parameter $\gamma$ (\ref{eqn:tanh_upd}), where we use (\ref{eqn:p_distractor_av}) and (\ref{eqn:p_signal_av}) to find the variance bound (Fig. \ref{fig:tau_info}B). The non-linear functions do not precisely match the contracting weights, but the time constant approximation still holds fairly well (Fig. \ref{fig:tau_info}C).

With the clipped-linear activation function and the HDC coding framework, we can find further details of the information theoretic properties of distributed random codes. The network activations can be bounded to a discrete set of states by the clipping parameter, $\kappa$. Thus, the total bits required to represent the network state in binary is $I_{storage} = N \log_2 ( 2 \kappa + 1 )$. 

The channel capacity of a network bound by the parameter $\kappa$ can be compared to the bits required for its binary representation. We first note that in general, the amount of information that can be retrieved from the network increases with higher $\kappa$ (Fig. \ref{fig:clipping_capacity}E1) and with larger $D$ (Fig. \ref{fig:clipping_capacity}F1). However, as $\kappa$ increases so too do the number of bits required to represent the network. There is a peak bit-per-bit maximum with 4-6 storage bits per neuron, dependent upon $D$ (Fig. \ref{fig:tau_info}D).

\section{Discussion}

%Recapitulate our main motivation
Leveraging a form of superposition is fundamental for computation in vector-symbolic architectures. Superposition is used to summarize sets or sequences of items in an ongoing computation. Superposition is a lossy compression of information and therefore VSA models have to be carefully designed to guarantee that the accuracy in retrieving items from a superposition is sufficient. Various VSA models have been proposed to describe cognitive reasoning and for artificial intelligence, but a general characterization and quantitative analysis of the used superposition principle was still lacking. We derived a theory for precisely describing the performance of superposition retrieval, and we developed superposition schemes for novel applications, such as for handling continuous data streams.

\subsection{Gaussian theory for VSA models mapped to RNNs}

We described an approach to map the encoding of symbols, sequence indexing and superposition in a given VSA model to an equivalent recurrent neural network. This approach not only suggests concrete neural implementations of proposed algorithms, but also allowed us to trace and highlight common properties of existing VSA models and develop a theory for analyzing them. 

Analyzing recurrent neural networks is extremely difficult in general, because complicated statistical dependencies arise from the recurrent feedback. However, there are classes of recurrent networks with randomized weights that are easier to analyze. We formulated conditions (\ref{eqn:phiiid})-(\ref{eqn:wind}) under which the error for retrieving a component from the recurrent network state can be estimated with Gaussian theory. We then derived the Gaussian theory, which is the first theory describing a broad variety of VSA models, and which covers all error regimes.
Interestingly, all VSA models described in previous literature that fulfill conditions (\ref{eqn:phiiid})-(\ref{eqn:wind}) follow the same theory.  The previous theories for specific VSA models \citep{Plate2003, Gallant2013} were limited to the high-fidelity regime. Our theory and the approximations we derive for the high-fidelity regime show that the crosstalk error is smaller than estimated by the previous work. 

The proposed theory reveals the precise impact of crosstalk. A few simple take-away messages result from our finding that, for large $M$, retrieval accuracy is just a function of the ratio between the numbers of neurons $N$ and superposed items $M$, specifically, the signal-to-noise ratio is $s=\sqrt{N/M}$ (\ref{eqn:p_corr_lin_largeM}). This implies that high-fidelity retrieval (i.e., negligible error correction) requires the number of superposed elements to be smaller than the number of neurons. Further, the accuracy is independent of the moments of a particular distribution of random codes, which explains why the performance of different VSA models is the same.

In addition, we generalized the Gaussian theory to neural networks with contracting weights or saturating nonlinear neurons. To our knowledge, the proposed analytic treatment for these types of neural networks is novel. These networks produce superpositions of the input sequence that include a recency or palimpsest effect. The recency effect leads to gradual weakening of symbols that have arrived many steps in the past, thereby creating space for newly arriving symbols and preventing catastrophic forgetting. Both mechanisms analyzed have the same qualitative effects and they can be used to construct \emph{memory buffers}, which memorize the recent past in continuous input streams without suffering from catastrophic forgetting. We used the theory to derive parameters for optimizing the information capacity and for controlling the trade-off between retrieval accuracy and memory span.

\subsection{Information capacity of superposition representations}
The channel capacity of superposition is the mutual information between the superposed symbols and the symbols estimated by retrieval. The novel theory allowed us to compute the channel capacity in all error regimes. We find that the superposition capacity scales linearly with the dimension $N$ of the representations, achieving a maximum of about $0.5$ bits per neuron, higher than predicted by previous theories. 
Our result that retrieval accuracy for large $M$ is a function of the ratio $N/M$ can explain the observation that the information capacity is $O(1)$. But given that a component in the hypervector posesses an entropy of order $O(\sqrt{M})$, a capacity of order $O(1)$ seems surprisingly low. We found that only in the case $M=1$ (that is, without superposition) entropy and capacity approach each other.
In the presence of superposition, the reduction in capacity is due to the fact that superposition is a form of lossy compression and redundancy is required to maintain high retrieval accuracy. Further, every item or group of items stored in memory is linearly separable, which sacrifices some memory capacity for easy computation.
For example, even when superposing just two code vectors ($M=2$), the entropy of $1.5$ bits per neuron exceeds the channel capacity, which is about $0.5$ bits per neuron (Fig. \ref{fig:linear_fit-info}F).
We found that the highest capacity values near $0.5$ bits per neuron are assumed for smaller numbers $M$ of superposed items. However, the decrease with growing $M$ is shallow (Fig. \ref{fig:linear_fit-info}E,F), keeping the capacity between $0.3-0.5$ bits per neuron. 

% Using learning to find good superposition schemes
Further, we asked to what extent the i.i.d. randomized coding in standard VSA models is optimized for efficient superposition. To address this question we investigated whether learning in a recurrent neural network can discover a (dependent) set of code vectors that produces higher capacity than independently drawn random vectors. Our results suggest that typical learning algorithms, such as backpropagation are unable to increase the capacity above values achieved by the traditional randomized coding schemes (Fig. \ref{fig:hd_capacity}G,H).

\subsection{Connections to the neural network literature} 
Our results for contracting matrices (Results \ref{sec:eigen_decay}) and for input noise (Results \ref{sec:noise}) describe the same types of networks considered in \citet{White2004} and \citet{Ganguli2008}. In agreement, we found that information content is maximized when the recurrent weights are unitary and the neurons are linear. They demonstrate optimized
``Distributed Shift Register'' models (DSR) which can fully exploit the entropy of binary neurons, that is, to achieve a channel capacity of 1 bit per neuron. 
These models correspond to our case of $M=1$, so they cannot serve to model superpositions, but are nevertheless instructive to consider (see Methods \ref{sec:dsr}). DSR models achieve 1 bit per neuron in finite-sized systems using a constructed codebook, not random code vectors. Our analysis of the $M=1$ case confirms that 1 bit per neuron is achievable with a constructed codebook (Methods \ref{sec:dsr}). For random codes, there is a small reduction of the capacity due to duplication of code vectors (Methods \ref{sec:random1}) -- for large network sizes this reduction becomes negligible. Also, compared to VSA models, DSR models break down catastrophically for small amounts of noise  (Methods Fig. \ref{fig:dsr_noise}). 

\citet{Ganguli2008} discussed in passing the effect of saturating non-linear activation functions and suggested a weakened scaling law of $O(\sqrt{N})$; hence a network not suitable for ``extensive memory'', which is defined as information maintenance with $O(N)$ scaling. However, our results illustrate that the information content in such saturating non-linear networks does scale with $O(N)$. This result requires taking into account the interdependence between noise (i.e. $\sigma_\eta$), network time constant (i.e. $\tau$), and number of neurons ($N$). Note, that the $O(\sqrt{N})$ in \citet{Ganguli2008} scaling law was derived for a fixed dynamic range of the activations ($\kappa$) as $N$ grows large, rather than considering the optimized dynamic range depending on $N$.
 
The presented analysis of superposition operations with recency effect is potentially important for other models which use network activity to form working memory \citep{Jaeger2002a, Maass2002} and to learn and generate complex output sequences \citep{Sussillo2009}. The derived buffer time constant $\tau$ and its relationship to network scale and code parameters can be used to optimize and understand network limits and operational time-scales.

\newpage

\section{Methods}

\subsection{Vector symbolic architectures}

\subsubsection{Basics}

\label{sec:vsa_basics}

The different vector symbolic architectures described here share many fundamental properties, but also have their unique flavors and potential advantages/disadvantages. Each framework utilizes random high-dimensional vectors (hypervectors) as the basis for representing symbols, but these vectors are drawn from different distributions. Further, different mechanisms are used to implement the key operations for vector computing: \emph{superpostion}, \emph{binding}, \emph{permutation}, and \emph{similarity}.

%& \text{\cite{MAP98}}\text{\cite{Plate2003}& \text{\cite{Plate2003}}& \text{\cite{Gallant}}& \text{\cite{Rachkovskij2001}}  & \text{\cite{GAHRR}} 

\begin{table}
\caption{Summary of VSA computing frameworks. Each framework has its
  own set of symbols and operations on them for addition,
  multiplication, and a measure of similarity.
  % BnSC: Bipolar Spatter Code, \citet{BSC96}; MAP: Multiply-Add-Permute \citet{MAP98}
  }
\label{tab:vsas}
\begin{equation*}
\begin{array}{lllll} %{p{1cm}p{1cm}p{1cm}p{1cm}p{1cm}p{1cm}}
\text{VSA} &  \text{Symbol Set} & \text{Binding}  & \text{Permutation} & \text{Trajectory Association} \\ \hline
%\text{BSC} & \text{dense binary} & \text{XOR} & \text{majority} & \rho & \text{Hamming} \\ 
\text{HDC} & \mathcal{B} := \{ -1, +1 \}^N & \times & \rho & \sum \rho^m (\mathbf{\Phi}_{d'}) \\ 
\text{HRR} & \mathcal{N}(0, 1/N)^N & \circledast  & \text{none} & \sum w^m \circledast \mathbf{\Phi}_{d'} \\ 
\text{FHRR} & \mathcal{C} := \{e^{i\mathcal{U}(0, 2\pi)} \}^N  & \times  & \circledast & \sum w^m \times \mathbf{\Phi}_{d'} \ \text{or} \  \sum w^m \circledast \mathbf{\Phi}_{d'} \\ 
%\text{MBAT} & \text{dense bipolar} & \text{vector-matrix} & + & \text{multiple} & \text{dot product} \\
   %    &  &  \text{multiplication}  &  & \text{binding} & \\
%\text{BSDC} & \text{sparse binary}   & \text{context-dependent} & \text{element-wise} & \rho & \text{overlap} \\
%      &   &   \text{thinning}   &  \text{disjunction}  & & \\
%\text{GAHRR} & \text{unit} & \text{geometric product} & + & \text{none} & \text{unitary-space} \\
% &  &  &  & & \text{scalar product} \\
\end{array}
\end{equation*}
\end{table}

The \emph{similarity} operation transforms two hypervectors into a scalar that represents similarity or distance. In HDC, HRR, FHRR and other frameworks, the similarity operation is the dot product of two hypervectors, while the Hamming distance is used in the frameworks which use only binary activations.  The distance metric is inversely related to the similarity metric. When vectors are similar, then their dot product will be very high, or their Hamming distance will be close to 0. When vectors are orthogonal, then their dot product is near 0 or their Hamming distance is near 0.5.

When the \emph{superposition} ($+$) operation is applied to a pair of hypervectors, then the result is a new hypervector that is similar to each one of the original pair. Consider HDC, given two hypervectors, $\mathbf{\Phi}_A, \mathbf{\Phi}_B$, which are independently chosen from $\mathcal{B} := \lbrace -1, +1 \rbrace ^ N$ and thus have low similarity ($\mathbf{\Phi}_A^\top \mathbf{\Phi}_B = 0 + noise$), then the superposition of these vectors, $\mathbf{x} := \mathbf{\Phi}_A + \mathbf{\Phi}_B$, has high similarity to each of the original hypervectors (e.g. $\mathbf{\Phi}_A^\top \mathbf{x} = N + noise$). In the linear VSA frameworks \citep{Kanerva2009}, \citet{Plate2003}, we do not constrain the \emph{superposition} operation to restrict the elements of the resulting vector to $\lbrace -1, +1 \rbrace$, but we allow any rational value. However, other frameworks \citep{Kanerva1996, Rachkovskij2001a} use clipping or majority-rule to constrain the activations, typically to binary values. 

The \emph{binding} operation ($\times$) combines two hypervectors into a third hypervector ($\mathbf{x} := \mathbf{\Phi}_A \times \mathbf{\Phi}_B$) that has low similarity to the original pair (e.g. $\mathbf{\Phi}_A^\top \mathbf{x}= 0 + noise$) and also maintains its basic statistical properties (i.e. it looks like a vector chosen from $\mathcal{B}$). In the HDC framework, the hypervectors are their own multiplicative self-inverses (e.g. $\mathbf{\Phi}_A \times \mathbf{\Phi}_A = \mathbf{1}$, where $\mathbf{1}$ is the binding identity), which means they can be ``dereferenced'' from the bound-pair by the same operation (e.g. $\mathbf{\Phi}_A \times \mathbf{x} = \mathbf{\Phi}_B + noise$). In the binary frameworks, the binding operation is element-wise \verb|XOR|, while in HRR and other frameworks binding is implemented by circular convolution ($\circledast$).
 
In different VSA frameworks, these compositional operations are implemented by different mechanisms. We note that all the binding operations can be mapped to a matrix multiply and the frameworks can be considered in the same neural network representation. The FHRR framework is the most generic of the VSAs and can utilize both multiply ($\times$) and circular convolution ($\circledast$) as a binding mechanism.

\subsubsection{Implementation details}
\label{sec:alt_vsa_analysis}

The experiments are all implemented in python as jupyter notebooks using standard packages, like numpy.

The experiments done with different VSA frameworks use different implementations for binding, most of which can be captured by a matrix multiplication. However, for efficiency reasons, we implemented the permutation operation $\rho$ and the circular convolution operation $\circledast$ with more efficient algorithms than the matrix multiplication. The permutation operation can be implemented with $O(N)$ complexity, using a circular shifting function (\verb|np.roll|). Efficient circular convolution can be performed by fast Fourier transform, element-wise multiply in the Fourier domain, and inverse fast fourier transform, with $O(N log N)$ complexity.

To implement FHRR, we utilized a network of dimension $N$, where the first $N/2$ elements of the network are the real part and the second $N/2$ elements are the imaginary part. Binding through complex multiplication is implemented as:
\begin{equation*}
\begin{split}
\mathbf{a} \times \mathbf{b} &= \left[ \begin{array}{c} 
\mathbf{a}_{real} \times \mathbf{b}_{real} - \mathbf{a}_{imaginary} \times \mathbf{b}_{imaginary} \\
\mathbf{a}_{real} \times \mathbf{b}_{imaginary} + \mathbf{a}_{imaginary} \times \mathbf{b}_{real} \\
\end{array} \right]  \\
%&= \left[ \begin{array}{c}
%A[:N/2] \times B[:N/2] - A[N/2:] \times B[N/2:] \\
%A[:N/2] \times B[N/2:] + A[N/2:] \times B[:N/2] \\
%\end{array} \right]         
\end{split}            
\end{equation*}

%The \emph{superposition} ($+$) and \emph{similarity} ($\cdot$) functions remain the same in this implementation. 
The \emph{circular convolution} operation can also be implemented in this framework, but with consideration that the pairs of numbers are permuted together. This can be implemented with a circulant matrix $\mathbf{W}$ with size $(N/2, N/2)$:
\begin{equation*}
\mathbf{w} \circledast \mathbf{a} = \left[ \begin{array}{cc}
\mathbf{W} & \mathbf{0} \\
\mathbf{0} & \mathbf{W} \\
\end{array} \right] \mathbf{a}
\end{equation*}

The \emph{superposition} ($+$) is the same, and \emph{similarity} ($\cdot$) functions is defined for complex numbers as simply:
\begin{equation*}
 \mathbf{a} \cdot \mathbf{b}  = \mathbf{a}_{real} \cdot \mathbf{b}_{real} + \mathbf{a}_{imaginary} \cdot \mathbf{b}_{imaginary}
 \end{equation*}
which is the real part of the conjugate dot product, $Re(\mathbf{a}^\top \mathbf{b}^*)$.
 
Either circular convolution or element-wise multiplication can be used to implement binding in FHRR, and trajectory association can be performed to encode the letter sequence with either operation:
\begin{equation*}
\begin{split}
\mathbf{x}(M) &= \sum \mathbf{w}^ {M-m} \circledast \mathbf{\Phi} \mathbf{a}(m) \  \text{or} \\ 
\mathbf{x}(M) &= \sum \mathbf{w}^{M-m} \times \mathbf{\Phi} \mathbf{a}(m)
\end{split}
\end{equation*}

\subsection{Training procedure for the recurrent neural network}
\label{sec:rnn_train}

We used tensorflow to train a linear recurrent neural network at the letter sequence recall task. The parameter $K$ could be altered to train the network to output the letter given to it in the sequence $K$ time steps in the history. The training was based on optimizing the Energy function given by the cross-entropy between $\mathbf{a}(m-K)$ and $\mathbf{\hat{a}}(m-K)$. 
The accuracy was monitored by comparing the maximum value of the output histogram with the maximum of the input histogram. 

We initialized the network to have a random Gaussian distributed encoding and decoding matrix ($\mathbf{\Phi}, \mathbf{V}_K$) and a fixed random unitary recurrent weight matrix ($\mathbf{W}$). The random unitary matrix was formed by taking the unitary matrix from a QR decomposition of a random Gaussian matrix. Such a matrix maintains the energy of the network, and with a streaming input, the energy of the network grows over time. After a fixed number of steps ($M=500$), the recurrent network was reset, where the activation of each neuron was set to 0. This erases the history of the input. Only outputs $K$ or more time steps after each reset were consider part of the energy function. 

\subsection{Accuracy of retrieval from superpositions}

\begin{figure}[t]
\begin{lstlisting}
from __future__ import division
import numpy as np
import scipy.special

def ncdf(z):
    return 0.5 * (1 + scipy.special.erf(z/2**0.5))

def p_correct_snr(M, N=10000, D=27, ares=2000):
    p = np.zeros((ares-1, len(M)))
    for iM,Mval in enumerate(M):
        s = (N / Mval)**0.5
        # span the Hit distribution up to 8 standard deviations
        av = np.linspace(s - 8, s + 8, ares)
        # the discretized gaussian of h_d'
        pdf_hdp = ncdf(av[1:])-ncdf(av[:-1])
        # the discretized cumulative gaussian of h_d 
        cdf_hd = ncdf(np.mean(np.vstack((av[1:]+s, av[:-1]+s)), axis=0))
        p[:, iM] = pdf_hdp * cdf_h ** (D-1)
    return np.sum(p, axis=0) # integrate over av
\end{lstlisting}
\caption[Numeric algorithm for $p_{corr}$]{\textbf{Numeric algorithm for accuracy integral.}}
\label{fig:p_corr_algo}
\end{figure}

\subsubsection{Comparsion of the different approximations for the high-fidelity regime}
\label{sec:lb_p_corr}

 \begin{figure}[t]
 \centering
 \includegraphics[width=\textwidth]{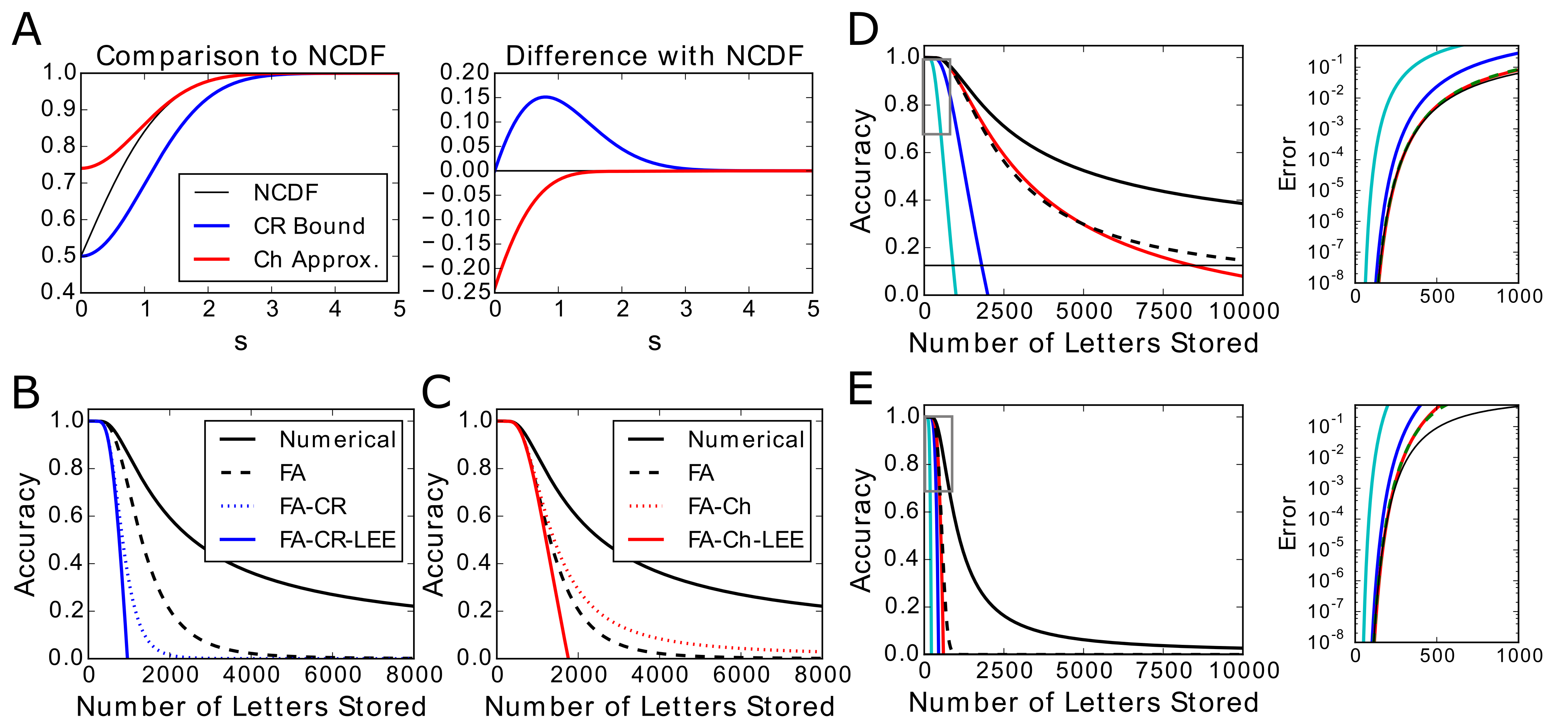}
 \caption[Comparison of approximation steps]{\textbf{Comparison of different methods to approximate the retrieval accuracy.} A. The Chernoff-Rubin (CR) \citep{chernoff1952measure} lower bound (blue) and the \citet{chang2011chernoff} approximation (red) to compute the normalized cumulative density function (NCDF; black) analytically. The \citet{chang2011chernoff} approximation becomes tight faster in the high-fidelity regime, but is not a lower bound. B. Differences between the three methods of approximations and the numerically evaluated $p_{corr}$ integral (black line). The factorial approximation (dashed black line) still requires numerical evalutation of the NCDF. Adding the CR lower-bound (dashed blue) and further the local-error expansion the high-fidelity regime can still be described well but the low-fidelity regime cannot be captured. C. Same as B, but using the \citet{chang2011chernoff} approximation to the NCDF. D. Accuracy curve and approximations for $D=8$. E. $D=1024$. Right panels in D and E are zoom in's into the high-fideltiy regime (marked by gray box insets in the left panels).}
 \label{fig:approx_compare}
 \end{figure}
 
 We compared each step of the approximation to the true numerically evaluated integral, to understand which regimes the approximations were valid (Fig. \ref{fig:approx_compare}B).

 We compare the CR bound and the \citet{chang2011chernoff} approximation to the numerically evaluated $\Phi$ and see that the CR lower bound does not get tight until multiple standard deviations into the very high-fidelity regime (Fig. \ref{fig:approx_compare}A). 
 
In Fig. \ref{fig:approx_compare}D, E, we see that while the approximations given are not strictly lower bounds, they are typically below the numerically evaluated accuracy. The Chang approximation can over-estimate the performance, however, in the high-fidelity regime when $D$ is large.

\subsubsection{Previous theories of the high-fidelity regime}
\label{sec:plate_compare}

The capacity theory derived here is similar to, but slightly different from the analysis of \citet{Plate2003}, which builds from work done in \citet{Plate1991}.
% \citet{Plate2003}'s framework for capacity analysis is set up differently, but can be roughly translated.
 \citet{Plate2003} frames the question: ``What is the probability that I can correctly decode all $M$ tokens stored, each of which are taken from the full set of $D$ possibilities without replacement?'' This is a slightly different problem, because this particular version of \cite{Plate2003}'s anlaysis does not use trajectory association to store copies of the same token in different addresses. Thus $M$ is always less than $D$, the $M$ tokens are all unique, and there is a difference in the sampling of the tokens between our analysis frameworks.

Nonetheless, these can be translated to a roughly equivalent framework given that $D$ is relatively large compared to $M$. \citet{Plate2003} derives the hit $p(\mathbf{h}_{d'})$ and reject $p(\mathbf{h}_d)$ distributions in the same manner as presented in our analysis, as well as uses a threshold to pose the probability problem:
\begin{equation}
p_{all-corr} = p(\mathbf{h}_{d'} > \theta)^M p(\mathbf{h}_d < \theta) ^{D-M}
\label{eqn:plate_pr_all}
\end{equation}

This can be interpreted as: the probability of reading all $M$ tokens correctly ($p_{corr-all}$) is the probability that the dot product of the true token is larger than threshold for all $M$ stored tokens ($p(\mathbf{h}_{d'} > \theta)^M$)  \emph{and} that the dot product is below threshold for all $D-M$ remaining distractor tokens ($p(\mathbf{h}_d < \theta)^{D-M}$). 

In our framework, the probability of correctly reading out an individual token from the $M$ tokens stored in memory is independent for all $M$ tokens. This is \ref{eqn:p_corr_s}, and to alter the equation to output the probability of reading all $M$ tokens correctly, then simply raise $p_{corr}$ to te $M$th power: 
\begin{equation}
p_{all-corr} = \left[ p_{corr} \right] ^ M = \left[  \int_{\theta}^{\infty} \frac{dh}{\sqrt{2 \pi}} e^{\frac{-h^2}{2}} \left[ \Phi\left( h + s \right) \right]^{D-1}    \right]^M
\end{equation}

\begin{figure}[t]
\centering
\includegraphics[width=0.7\textwidth]{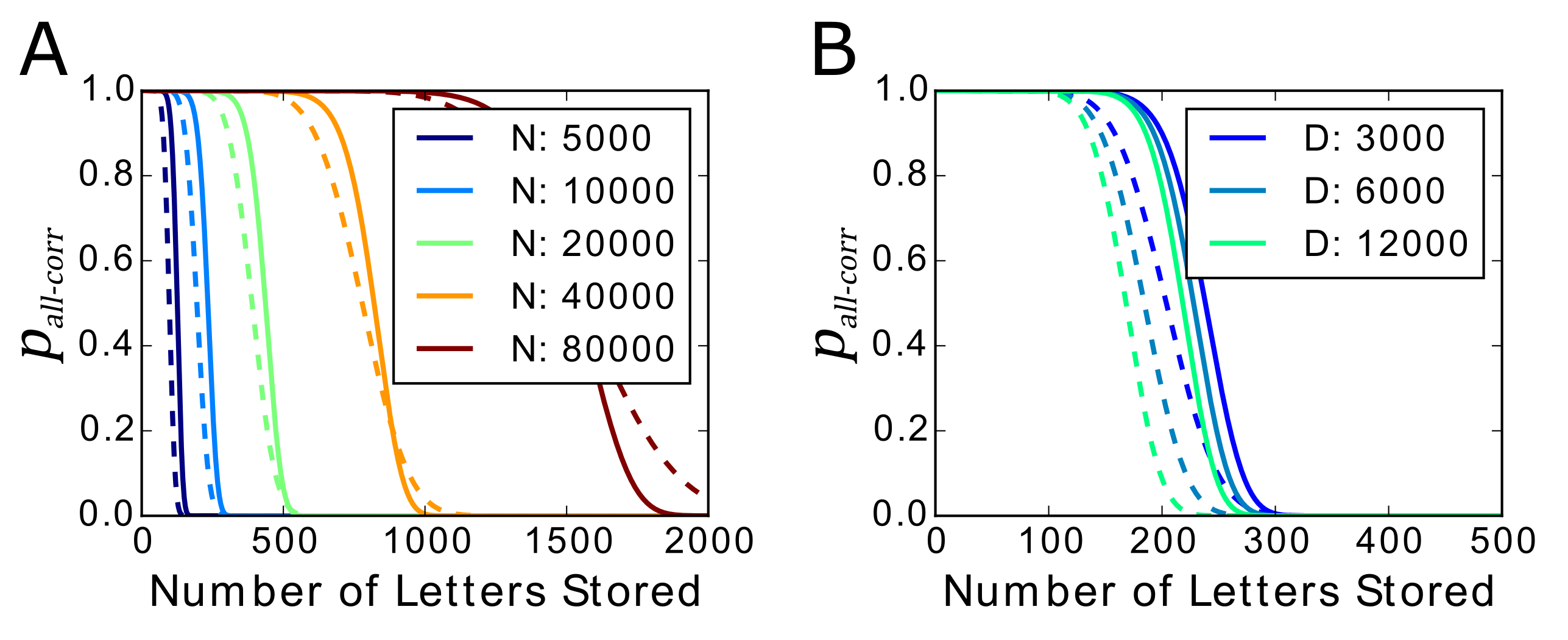}
\caption[Comparison with the theory in \citet{Plate2003}]{\textbf{Comparison with the theory in \citet{Plate2003}.}
A. \citet{Plate2003} derived $p_{all-corr} = p_{corr}^M$, plotted in dashed lines for different values of $N$ with $D$ fixed at 4096. The new theory in solid lines.
B. \citet{Plate2003}'s theory in dashed lines with different values of $D$ and fixed $N$. The new theory in solid lines.
}
\label{fig:plate_comparison}
\end{figure}

In Figure \ref{fig:plate_comparison}, we compare our theory to Plate's by computing $p_{all-corr}$ given various different parameters of $N$, $M$, and $D$. We show that \citet{Plate2003}'s framework comparatively underestimates the capacity of hypervectors. There is slight discrepancy in our analysis frameworks, because of how the tokens are decoded from memory. In our analysis framework, we take the maximum dot product as the decoded letter, and there are instances that can be correctly classified that \citet{Plate2003}'s probability statement (\ref{eqn:plate_pr_all}) would consider as incorrect. For instance, the true token and a distractor token can both have dot products above threshold and the correct token can still be decoded as long as the true token's dot product is larger than the distractor token. However, this scenario would be classified as incorrect by (\ref{eqn:plate_pr_all}).

Plate next uses an approximation to derive a linear relationship describing the accuracy. Citing \citet{Abramowitz1965}, he writes:
$$
erfc(x) < \frac{1}{x\sqrt{\pi}} e^{-x^2}
$$

This approximation allows Plate to estimate the linear relationship between $N$, $M$, $\log D$, and $\epsilon$. Arriving at:
$$
N < 8 M \log \left( \frac{D}{\epsilon} \right)
$$  

The FA-CR-LEE approximation only differs by a factor of 2, because of the slightly different choice we made to approximate the cumulative Gaussian as well as the different set-up for the problem.

Subsequent work by \citet{Gallant2013} proposed an alternative VSA framework, which used a matrix as a binding mechanism. Based on their framework, they too in analogous fashion to \citet{Plate2003} derived an approximation to the capacity of vector-symbols in superposition. Their derivation takes the high-fidelity factorial approximation as the starting point, and utilizes $e^{-x}$ as the bound on the tail of the normal distribution. This work is very similar to the derivation presented in this paper, but we add more rigor and derive a tighter high-fidelity approximation utilizing the Chernoff-Rubin bound and the updated approximation by \citet{chang2011chernoff}.

\subsection{Distributed shift registers: vector representations without superposition}
\label{sec:M1}

\subsubsection{Constructed vector representations}
\label{sec:dsr}

\citet{White2004} describes the distributed shift register (DSR), a neural network which encodes a binary sequence with a constructed codebook rather than using randomly drawn code vectors. Extended by \citet{Ganguli2008}, they use the Fisher Memory Curve (FMC) to study the capacity limitations of recurrent neural networks using the DSR, and derive a result showing that the capacity of neural networks is at most $N$ bits per neuron.
The FMC is an approximate measure of the memory of the network and is related to $\mathbf{h}_d$ and $s$ derived here. The DSR is a special case of a constructed representation, which does not utilize any superposition properties. Rather, the DSR is a construction process that builds a binary representation for each possible sequence and only stores a single binary representation in the network.

Here, we describe an extension of the DSR that can maintain 1 bit per neuron retrieval capacity, but these frameworks lack superposition and lose many of the advantages of distributed representations, such as tolerance to noise. In the DSR, a binary code is used to represent each of the $D=2$ tokens, with a permutation used to encode sequence position. \citet{White2004} and \citet{Ganguli2008} only consider this code when $D=2$. For $D>2$, the framework can be continued to maintain 1 bit per neuron as follows. Consider $D=16$, to create a DSR one would then enumerate the $D=16$ binary sequences of length $\log_2 D=4$ to represent each token. Each vector in the codebook $\mathbf{\Phi}$, would then contain a sequence of $+1$ or $-1$ with length  $\log_2 D=4$, followed by $0$s. The sequence of tokens would then be stored in the network using a permutation matrix that rotates the codeword $\log_2 D=4$ positions. This creates a binary code that utilizes all $2^N$ possible codewords to represent all sequences and stores $N$ bits of information without any crosstalk. However, it is clear that the codewords are not orthogonal to each other, and the code is not represented in superposition but rather in concatenation. Such a constructed representation can utilize the full bit per neuron entropy, but loses the benefits of distributed representations.

\begin{figure}
\centering
\includegraphics[width=0.8\textwidth]{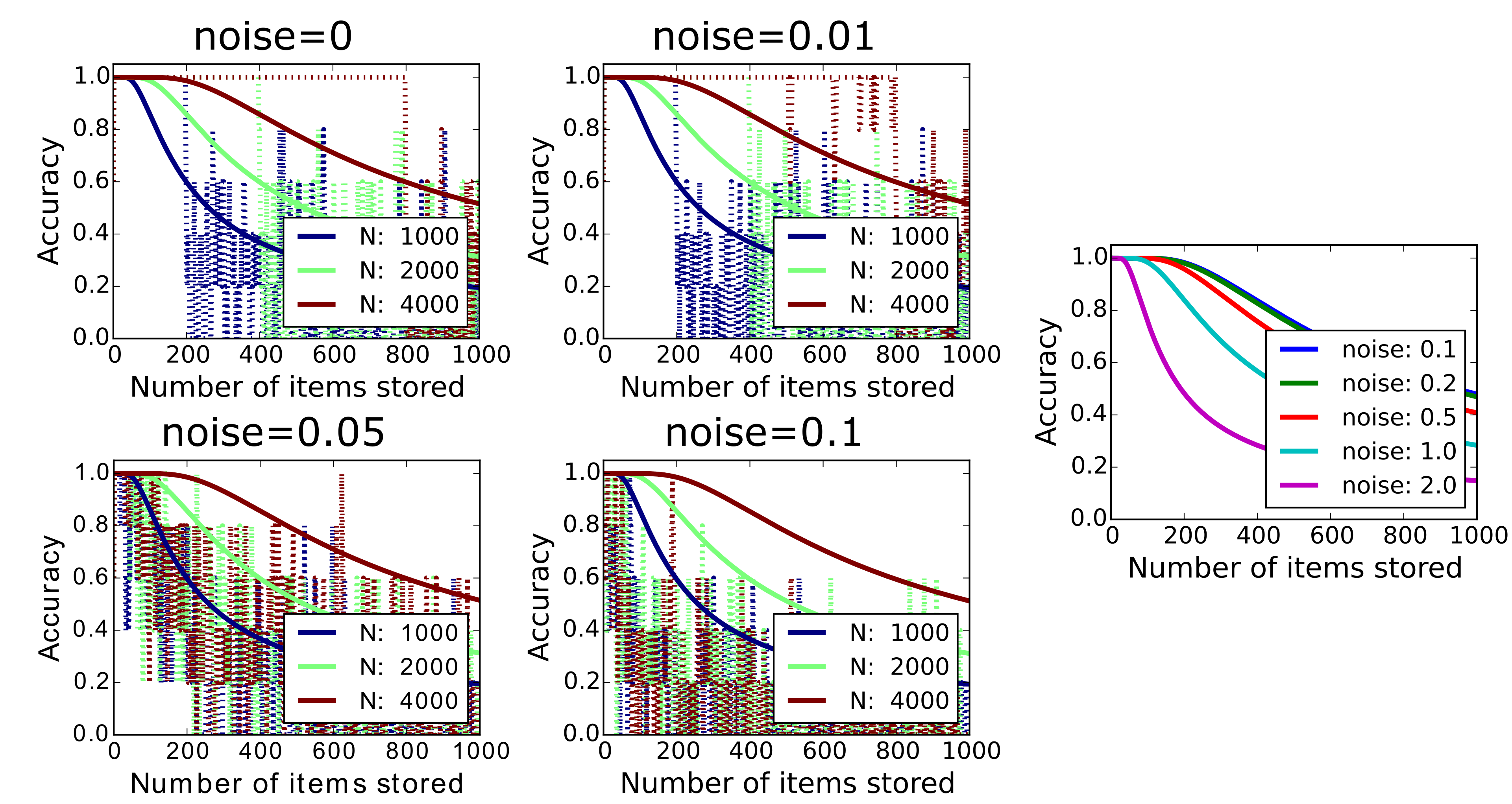}
\caption[DSR noise]{\textbf{DSR with constructed codes versus randomized codes.} Accuracy as a function of the dictionary size. With low noise the DSR (dashed lines) can fully exploit the entropy of binary representations and achieve a capacity of 1-bit per neuron (upper left panel). However, as noise slightly increases, the capacity of DSR's drops quickly. Superposition codes show robustness to noise (solid lines; right panel).}
\label{fig:dsr_noise}
\end{figure}

We illustrate the DSR scheme in Fig. \ref{fig:dsr_noise}. Compared with superposition codes, the DSR maintains the sequence perfectly until catastrophic collapse when $M \log_2 D > N$. However, since this coding scheme does not rely on distributed superposition, it does not develop noise tolerance as VSA codes are known to do \citep{Plate2003}. For relatively small amounts of noise the DSR coding scheme fails completely, while random codes can compensate for noise arbitrarily by increasing $N$ (Fig. \ref{fig:dsr_noise}).

\subsubsection{Randomized vector representations}
\label{sec:random1}
\begin{figure}[t]
\centering
\includegraphics[width=0.7\textwidth]{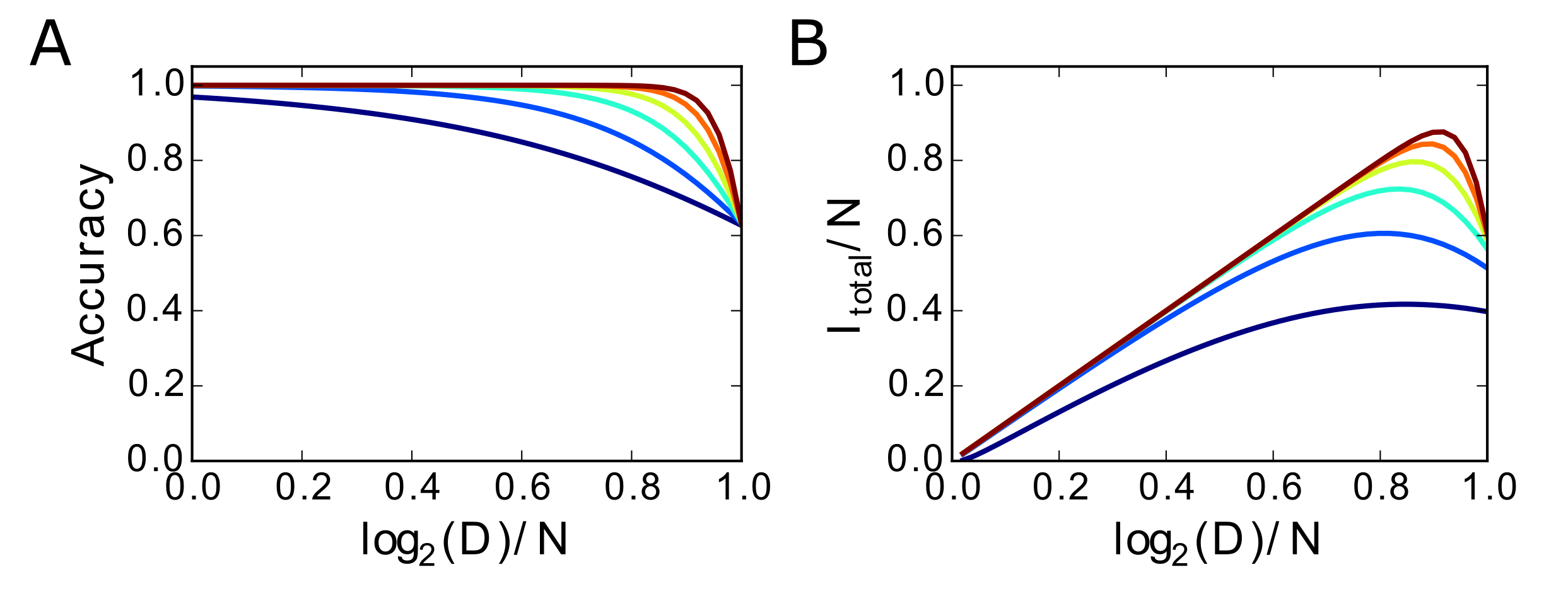}
\caption[Finite size effects]{\textbf{Finite size effects on information capacity in DSR's with randomized codes.} A. The accuracy $p_{corr}^{M=1}$ with increasing $N$. B. The retrieved information with increasing $N$.}
\label{fig:largeD}
\end{figure}

In Results \ref{sec:linear_scaling}, we compared the channel capacity of superpositions to the channel capacity of the $M=1$ case as $D \to 2^N$. %The latter case is challenging to numerically evaluate accurately for large $N$. 
As $D$ grows to a significant fraction of $2^N$, the cross talk from superposition becomes overwhelming and the channel capacity is maximized for $M=1$. The retrieval errors are then only due to collisions between the randomized codevectors and the accuracy $p_{corr}^{M=1}$ is given by (\ref{eqn:p_corr_M1}).  
Fig. \ref{fig:largeD}A, shows the accuracy for $M=1$ as $D$ grows to $2^N$ with a randomly constructed codebook for different (smaller) values of $N$ -- for large $N$ the numerical evaluation of (\ref{eqn:p_corr_M1}) is difficult. As $N$ grows, the accuracy remains perfect for an increasingly large fraction of the $2^N$ possible code vectors. However, at the point $D=2^N$ the accuracy falls off to $(1-1/e)$, but this fall-off is sharper is $N$ grows larger. The information retrieved from the network also grows closer and closer to 1 bit per neuron as $N$ grows larger with $M=1$ (Fig. \ref{fig:largeD}B).

In Fig. \ref{fig:largeD}B the capacity $I_{total}/N$ of the randomly constructed codebook for $M=1$ was computed with the equation we developed for superposed codes (\ref{eqn:total_infoK}). However, the nature of the retrieval errors is different for $M=1$, rather than cross talk, collisions of code vectors is the error source. By performing an exhaustive analysis of the collision structure of a particular random codebook, the error correction can be limited to actual collisions and the capacity of such a retrieval procedure is higher. The information transmitted when using the full knowledge of the collision structure is:
\begin{equation}
I_{total} = \sum_c p_c \log_2 \left( \frac{p_c D} {c+1} \right)
\end{equation}

For $D = 2^N$ and $N \to \infty$, the total information of a random vector symbol approaches 1 bit per neuron:
\begin{equation}
\lim_{N \to \infty} \frac{1}{N} \sum_c p_c \left( N + \log_2 \left( \frac{p_c}{c+1} \right) \right) \to 1
\end{equation}

It is an interesting and somewhat surprising result in the context of DSRs that a random codebook yields asymptotically, for large $N$, the same capacity as a codebook in which collisions are eliminated by construction \citep{White2004}. But it has to be emphasized that a retrieval procedure, which uses the collision structure of the random codebook, is only necessary and advantageous for the $M=1$ case. For superpositions, even with just two code vectors ($M=2$), the alphabet size $D$ has to be drastically reduced to keep cross talk under control and the probability of collisions between random code vectors becomes negligible.

\section*{Acknowledgements}
The authors would like to thank Pentti Kanerva, Bruno Olshausen, Guy Isely, Yubei Chen, Alex Anderson, Eric Weiss and the Redwood Center for Thoeretical Neuroscience for helpful discussions and contributions to the development of this work. This work was supported by the Intel Corporation (ISRA on Neuromorphic architectures for Mainstream Computing).


{\small\begin{thebibliography}{35}
\providecommand{\natexlab}[1]{#1}
\providecommand{\url}[1]{\texttt{#1}}
\expandafter\ifx\csname urlstyle\endcsname\relax
  \providecommand{\doi}[1]{doi: #1}\else
  \providecommand{\doi}{doi: \begingroup \urlstyle{rm}\Url}\fi

\bibitem[Abramowitz et~al.(1965)Abramowitz, Stegun, and Miller]{Abramowitz1965}
M.~Abramowitz, I.~A. Stegun, and D.~Miller.
\newblock {Handbook of Mathematical Functions With Formulas, Graphs and
  Mathematical Tables (National Bureau of Standards Applied Mathematics Series
  No. 55)}, 1965.
%\newblock ISSN 00218936.

\bibitem[Chang et~al.(2011)Chang, Cosman, and Milstein]{chang2011chernoff}
S.~H. Chang, P.~C. Cosman, and L.~B. Milstein.
\newblock {Chernoff-type bounds for the Gaussian error function}.
\newblock \emph{IEEE Transactions on Communications}, 59\penalty0
  (11):\penalty0 2939--2944, 2011.

\bibitem[Chernoff(1952)]{chernoff1952measure}
H.~Chernoff.
\newblock {A measure of asymptotic efficiency for tests of a hypothesis based
  on the sum of observations}.
\newblock \emph{The Annals of Mathematical Statistics}, pages 493--507, 1952.

\bibitem[Chiani et~al.(2003)Chiani, Dardari, and Simon]{chiani2003new}
M.~Chiani, D.~Dardari, and M.~K. Simon.
\newblock {New exponential bounds and approximations for the computation of
  error probability in fading channels}.
\newblock \emph{IEEE Transactions on Wireless Communications}, 2\penalty0
  (4):\penalty0 840--845, 2003.

\bibitem[Danihelka et~al.(2016)Danihelka, Wayne, Uria, Kalchbrenner, and
  Graves]{Danihelka2016}
I.~Danihelka, G.~Wayne, B.~Uria, N.~Kalchbrenner, and A.~Graves.
\newblock {Associative Long Short-Term Memory}.
\newblock \emph{ArXiv}
\newblock 2016.

\bibitem[Eliasmith et~al.(2012)Eliasmith, Stewart, Choo, Bekolay, DeWolf, Tang,
  Tang, and Rasmussen]{Eliasmith2012}
C.~Eliasmith, T.~C. Stewart, X.~Choo, T.~Bekolay, T.~DeWolf, Y.~Tang, C.~Tang,
  and D.~Rasmussen.
\newblock {A large-scale model of the functioning brain.}
\newblock \emph{Science (New York, N.Y.)}, 338\penalty0 (6111):\penalty0
  1202--5, 2012.
%\newblock ISSN 1095-9203.
%\newblock \doi{10.1126/science.1225266}.

\bibitem[Feinstein(1954)]{Feinstein1954}
A.~Feinstein.
\newblock {A new basic theorem of information theory}.
\newblock \emph{Transactions of the IRE Professional Group on Information
  Theory}, 4\penalty0 (4):\penalty0 2--22, 1954.
%\newblock ISSN 2168-2690.
%\newblock \doi{10.1109/TIT.1954.1057459}.

\bibitem[Gallant and Okaywe(2013)]{Gallant2013}
S.~I. Gallant and T.~W. Okaywe.
\newblock {Representing Objects, Relations, and Sequences}.
\newblock \emph{Neural Computation}, 25\penalty0 (8):\penalty0 2038--2078,
  2013.
%\newblock ISSN 0899-7667.
%\newblock \doi{10.1162/NECO_a_00467}.

\bibitem[Ganguli and Sompolinsky(2010)]{Ganguli2010}
S.~Ganguli and H.~Sompolinsky.
\newblock {Short-term memory in neuronal networks through dynamical compressed
  sensing.}
\newblock \emph{Nips}, 2010.
%\newblock ISSN 00944467.
%\newblock \doi{10.1163/187633309X12563839996540}.

\bibitem[Ganguli et~al.(2008)Ganguli, Huh, and Sompolinsky]{Ganguli2008}
S.~Ganguli, B.~D. Huh, and H.~Sompolinsky.
\newblock {Memory traces in dynamical systems}.
\newblock \emph{Proceedings of the National Academy of Sciences}, 105\penalty0
  (48):\penalty0 18970--18975, 2008.
%\newblock ISSN 1091-6490.
%\newblock \doi{10.1073/pnas.0804451105}.
%\newblock URL
 % \url{http://www.pnas.org/cgi/content/abstract/105/48/18970{\%}255Cnpapers2://publication/uuid/AEEC32EE-1E9C-4D78-A3B2-C55B4EA1E0B0
 % http://www.pnas.org/cgi/content/abstract/105/48/18970{\$}{\%}255C{\$}npapers2://publication/uuid/AEEC32EE-1E9C-4D78-A3B2-C55B4EA1E0B0}.

\bibitem[Gayler(1998)]{Gayler1998}
R.~W. Gayler.
\newblock {Multiplicative binding, representation operators {\&} analogy}.
\newblock In \emph{Gentner, D., Holyoak, K. J., Kokinov, B. N. (Eds.), Advances
  in analogy research: Integration of theory and data from the cognitive,
  computational, and neural sciences}, New Bulgarian University,
  Sofia, Bulgaria, 1998.

\bibitem[Gayler(2003)]{Gayler2003}
R.~W. Gayler.
\newblock {Vector Symbolic Architectures answer Jackendoff's challenges for
  cognitive neuroscience}.
\newblock \emph{Proceedings of the ICCS/ASCS International Conference on
  Cognitive Science}, \penalty0 (2002):\penalty0 6, 2003.
%\newblock ISSN 0140-525X.
%\newblock \doi{10.1017/S0140525X06309028}.
%\newblock URL \url{http://arxiv.org/abs/cs/0412059}.

\bibitem[Graves et~al.(2014)Graves, Wayne, and Danihelka]{Graves2014}
A.~Graves, G.~Wayne, and I.~Danihelka.
\newblock {Neural Turing Machines}.
\newblock \emph{ArXiv}
\newblock 2014.
%\newblock ISSN 2041-1723.
%\newblock \doi{10.3389/neuro.12.006.2007}.
%\newblock URL \url{http://arxiv.org/abs/1410.5401}.

\bibitem[Graves et~al.(2016)Graves, Wayne, Reynolds, Harley, Danihelka,
  Grabska-Barwi{\'{n}}ska, Colmenarejo, Grefenstette, Ramalho, Agapiou, Badia,
  Hermann, Zwols, Ostrovski, Cain, King, Summerfield, Blunsom, Kavukcuoglu, and
  Hassabis]{Graves2016}
A.~Graves, G.~Wayne, M.~Reynolds, T.~Harley, I.~Danihelka,
  A.~Grabska-Barwi{\'{n}}ska, S.~G. Colmenarejo, E.~Grefenstette, T.~Ramalho,
  J.~Agapiou, A.~P. Badia, K.~M. Hermann, Y.~Zwols, G.~Ostrovski, A.~Cain,
  H.~King, C.~Summerfield, P.~Blunsom, K.~Kavukcuoglu, and D.~Hassabis.
\newblock {Hybrid computing using a neural network with dynamic external
  memory}.
\newblock \emph{Nature}, 538\penalty0 (7626):\penalty0 471--476, 2016.
%\newblock ISSN 0028-0836.
%\newblock \doi{10.1038/nature20101}.
%\newblock URL \url{http://arxiv.org/abs/1610.04211
%  http://www.nature.com/doifinder/10.1038/nature20101}.

\bibitem[Hellman and Raviv(1970)]{bound1970probability}
M.~E. Hellman and J.~Raviv.
\newblock {Probability of error, equivocation, and the Chernoff bound}.
\newblock \emph{IEEE Transactions on Information Theory}, 16\penalty0 (4),
  1970.

\bibitem[Jacobs(1966)]{jacobs1966probability}
I.~Jacobs.
\newblock {Probability-of-error bounds for binary transmission on the slowly
  fading Rician channel}.
\newblock \emph{IEEE Transactions on Information Theory}, 12\penalty0
  (4):\penalty0 431--441, 1966.

\bibitem[Jaeger(2002)]{Jaeger2002a}
H.~Jaeger.
\newblock {Short term memory in echo state networks}.
\newblock \emph{GMD Report 152}, 2002.
%\newblock URL
%  \url{papers://78a99879-71e7-4c85-9127-d29c2b4b416b/Paper/p14153{\$}{\%}255C{\$}nhttp://neuron-ai.tuke.sk/{\%}257B{~}{\%}257Dbundzel/diploma{\%}257B{\_}{\%}257Dtheses{\%}257B{\_}{\%}257Dstudents/2006/Martin
%  Sramko- Echo State NN in Prediction/STMEchoStatesTechRep.pdf
%  papers://78a99879-71e7-4c85-9127-d29c2b4b4}.

\bibitem[Kanerva(1996)]{Kanerva1996}
P.~Kanerva.
\newblock {Binary spatter-coding of ordered K-tuples}.
\newblock \emph{Lecture Notes in Computer Science (including subseries Lecture
  Notes in Artificial Intelligence and Lecture Notes in Bioinformatics)}, 1112
  LNCS:\penalty0 869--873, 1996.
%\newblock ISSN 16113349.
%\newblock \doi{10.1007/3-540-61510-5_146}.

\bibitem[Kanerva(2009)]{Kanerva2009}
P.~Kanerva.
\newblock {Hyperdimensional computing: An introduction to computing in
  distributed representation with high-dimensional random vectors}.
\newblock \emph{Cognitive Computation}, 1:\penalty0 139--159, 2009.
%\newblock ISSN 18669956.
%\newblock \doi{10.1007/s12559-009-9009-8}.

\bibitem[Kleyko and Osipov(2014)]{Kleyko2014}
D.~Kleyko and E.~Osipov.
\newblock {On Bidirectional Transitions between Localist and Distributed
  Representations: The Case of Common Substrings Search Using Vector Symbolic
  Architecture}.
\newblock \emph{Procedia Computer Science}, 41\penalty0 (C):\penalty0 104--113,
  2014.
%\newblock ISSN 18770509.
%\newblock \doi{10.1016/j.procs.2014.11.091}.
%\newblock URL \url{http://dx.doi.org/10.1016/j.procs.2014.11.091
%  http://linkinghub.elsevier.com/retrieve/pii/S1877050914015361}.

\bibitem[Luko{\v{s}}evi{\v{c}}ius and Jaeger(2009)]{Lukosevicius2009}
M.~Luko{\v{s}}evi{\v{c}}ius and H.~Jaeger.
\newblock {Reservoir computing approaches to recurrent neural network
  training}.
\newblock \emph{Computer Science Review}, 3\penalty0 (3):\penalty0 127--149,
  2009.
%\newblock ISSN 15740137.
%\newblock \doi{10.1016/j.cosrev.2009.03.005}.
%\newblock URL
%  \url{http://linkinghub.elsevier.com/retrieve/pii/S1574013709000173}.

\bibitem[Maass et~al.(2002)Maass, Natschl{\"{a}}ger, and Markram]{Maass2002}
W.~Maass, T.~Natschl{\"{a}}ger, and H.~Markram.
\newblock {Real-Time Computing Without Stable States: A New Framework for
  Neural Computation Based on Perturbations}.
\newblock \emph{Neural Computation}, 14\penalty0 (11):\penalty0 2531--2560,
  2002.
%\newblock ISSN 0899-7667.
%\newblock \doi{10.1162/089976602760407955}.
%\newblock URL
%  \url{http://www.mitpressjournals.org/doi/abs/10.1162/089976602760407955}.

\bibitem[Owen(1980)]{owen1980table}
D.~B. Owen.
\newblock {A table of normal integrals: A table}.
\newblock \emph{Communications in Statistics-Simulation and Computation},
  9\penalty0 (4):\penalty0 389--419, 1980.

\bibitem[Papoulis(1984)]{Papoulis1984}
A.~Papoulis.
\newblock {Probability, Random Variables, and Stochastic Processes}, 1984.

\bibitem[Plate(1991)]{Plate1991}
T.~A. Plate.
\newblock {Holographic Reduced Representations : Convolution Algebra for
  Compositional Distributed Representations}.
\newblock \emph{Proceedings of the 12th international joint conference on
  Artificial intelligence}, pages 30--35, 1991.
%\newblock URL
%  \url{http://citeseer.ist.psu.edu/viewdoc/summary?doi=10.1.1.33.1585}.

\bibitem[Plate(1993)]{Plate1993}
T.~A. Plate.
\newblock {Holographic Recurrent Networks}.
\newblock \emph{Advances in Neural Information Processing Systems 5}, \penalty0
  (1):\penalty0 34--41, 1993.
%\newblock URL \url{ftp://ftp.ci.tuwien.ac.at/pub/texmf/bibtex/nips-5.bib}.

\bibitem[Plate(2003)]{Plate2003}
T.~A. Plate.
\newblock \emph{{Holographic Reduced Representation: Distributed Representation
  for Cognitive Structures}}.
\newblock Stanford: CSLI Publications, 2003.
\newblock ISBN 978-1575864303.
%\newblock URL
%  \url{http://www.amazon.com/Holographic-Reduced-Representation-Distributed-Structures/dp/1575864304}.

\bibitem[Plate(1995)]{Plate1995}
T.~A. Plate.
\newblock {Holographic reduced representations}.
\newblock \emph{IEEE Transactions on Neural Networks}, 6\penalty0 (3):\penalty0
  623--641, 1995.
%\newblock ISSN 10459227.
%\newblock \doi{10.1109/72.377968}.
%\newblock URL \url{http://www.ncbi.nlm.nih.gov/pubmed/18263348
%  http://ieeexplore.ieee.org/document/377968/}.

\bibitem[Rachkovskij(2001)]{Rachkovskij2001}
D.~A. Rachkovskij.
\newblock {Representation and processing of structures with binary sparse
  distributed codes}.
\newblock \emph{IEEE Transactions on Knowledge and Data Engineering},
  13\penalty0 (2):\penalty0 261--276, 2001.
%\newblock ISSN 10414347.
%\newblock \doi{10.1109/69.917565}.

\bibitem[Rachkovskij and Kussul(2001)]{Rachkovskij2001a}
D.~A. Rachkovskij and E.~M. Kussul.
\newblock {Binding and Normalization of Binary Sparse Distributed
  Representations by Context-Dependent Thinning}.
\newblock \emph{Neural Computation}, 13\penalty0 (2):\penalty0 411--452, 2001.
%\newblock ISSN 0899-7667.
%\newblock \doi{10.1162/089976601300014592}.
%\newblock URL
%  \url{http://www.mitpressjournals.org/doi/abs/10.1162/089976601300014592}.

\bibitem[Rasmussen and Eliasmith(2011)]{Rasmussen2011}
D.~Rasmussen and C.~Eliasmith.
\newblock {A neural model of rule generation in inductive reasoning}.
\newblock \emph{Topics in Cognitive Science}, 3\penalty0 (1):\penalty0
  140--153, 2011.
%\newblock ISSN 17568757.
%\newblock \doi{10.1111/j.1756-8765.2010.01127.x}.

\bibitem[Rinkus(2012)]{Rinkus2012}
G.~J. Rinkus.
\newblock {Quantum computation via sparse distributed representation}.
\newblock \emph{NeuroQuantology}, 10\penalty0 (2):\penalty0 311--315, 2012.
%\newblock ISSN 13035150.

\bibitem[Sussillo and Abbott(2009)]{Sussillo2009}
D.~Sussillo and L.~F. Abbott.
\newblock {Generating coherent patterns of activity from chaotic neural
  networks.}
\newblock \emph{Neuron}, 63\penalty0 (4):\penalty0 544--57, 2009.
%\newblock ISSN 1097-4199.
%\newblock \doi{10.1016/j.neuron.2009.07.018}.
%\newblock URL
%  \url{http://www.pubmedcentral.nih.gov/articlerender.fcgi?artid=2756108{\&}tool=pmcentrez{\&}rendertype=abstract}.

\bibitem[White et~al.(2004)White, Lee, and Sompolinsky]{White2004}
O.~L. White, D.~D. Lee, and H.~Sompolinsky.
\newblock {Short-term memory in orthogonal neural networks}.
\newblock \emph{Physical Review Letters}, 92\penalty0 (14):\penalty0 148102--1,
  2004.
%\newblock ISSN 00319007.
%\newblock \doi{10.1103/PhysRevLett.92.148102}.

\bibitem[Williams and Zipser(1989)]{Williams1989cs}
R.~J. Williams and D.~Zipser.
\newblock {Experimental Analysis of the Real-time Recurrent Learning
  Algorithm}.
\newblock \emph{Connection Science}, 1\penalty0 (1):\penalty0 87--111,
  1989.
%\newblock ISSN 0954-0091.
%\newblock \doi{10.1080/09540098908915631}.
%\newblock URL \url{http://dx.doi.org/10.1080/09540098908915631
%  https://www.tandfonline.com/doi/full/10.1080/09540098908915631}.

\end{thebibliography}

}
\end{document}